\documentclass[letterpaper, 10 pt, conference]{ieeeconf}  

\IEEEoverridecommandlockouts                              

\usepackage{graphicx} 
\usepackage{amsmath} 
\usepackage{microtype}
\usepackage{siunitx}
\usepackage{units}
\usepackage{booktabs}
\usepackage{algorithm, algorithmic}
\usepackage{mathtools}

\let\labelindent\relax
\usepackage{enumitem}
\usepackage[hidelinks]{hyperref}

\usepackage{tikz}
\usepackage{pgfplots}
\usetikzlibrary{arrows,decorations,backgrounds,shapes,chains}
\pgfplotsset{compat=newest}
\pgfplotsset{plot coordinates/math parser=false}
\newlength\figureheight
\newlength\figurewidth

\usepackage[caption=false]{subfig}  

\usepackage{tabularx}

\makeatletter
\newcommand\footnoteref[1]{\protected@xdef\@thefnmark{\ref{#1}}\@footnotemark}
\makeatother

\newcommand{\footlabel}[2]{%
    \addtocounter{footnote}{1}%
    \footnotetext[\thefootnote]{%
        \addtocounter{footnote}{-1}%
        \refstepcounter{footnote}\label{#1}%
        #2%
    }%
    $^{\ref{#1}}$%
}

\newcommand{\footref}[1]{%
    $^{\ref{#1}}$%
}

\newcommand{\diag}{{\mathop{\mathrm{diag}}}}

\newcommand{\vth}{\mbox{\boldmath $\theta$}}

\newcommand{\vph}{\mbox{\boldmath $\varphi$}}

\newcommand{\vom}{\mbox{\boldmath $\omega$}}

\newcommand{\ve}{\mathbf e}
\newcommand{\vf}{\mathbf f}
\newcommand{\vg}{\mathbf g}
\newcommand{\vh}{\mathbf h}

\newcommand{\vl}{\mathbf l}

\newcommand{\vq}{\mathbf q}
\newcommand{\vr}{\mathbf r}
\newcommand{\vs}{\mathbf s}

\newcommand{\vu}{\mathbf u}
\newcommand{\vv}{\mathbf v}

\newcommand{\vx}{\mathbf x}

\newcommand{\vA}{\mathbf A}
\newcommand{\vB}{\mathbf B}
\newcommand{\vC}{\mathbf C}
\newcommand{\vD}{\mathbf D}

\newcommand{\vF}{\mathbf F}

\newcommand{\vI}{\mathbf I}
\newcommand{\vJ}{\mathbf J}
\newcommand{\vK}{\mathbf K}
\newcommand{\vL}{\mathbf L}

\newcommand{\vP}{\mathbf P}
\newcommand{\vQ}{\mathbf Q}
\newcommand{\vR}{\mathbf R}
\newcommand{\vS}{\mathbf S}

\title{\LARGE \bf
Efficient Kinematic Planning for Mobile Manipulators with Non-holonomic Constraints Using Optimal Control}

\author{Markus Giftthaler, Farbod Farshidian, Timothy Sandy, Lukas Stadelmann and Jonas Buchli$^\ast$
\thanks{$^*$Agile \& Dexterous Robotics Lab, Institute of Robotics and Intelligent Systems, ETH Z\"urich, Switzerland. {\small \{mgiftthaler, farshidian, tsandy, stalukas, buchlij\}@ethz.ch}}
}

\begin{document}
\maketitle
\thispagestyle{empty}
\pagestyle{empty}

\begin{abstract}
This work addresses the problem of kinematic trajectory planning for mobile manipulators with non-holonomic constraints, and holonomic operational-space tracking constraints. We  obtain whole-body trajectories and time-varying kinematic feedback controllers by solving a Constrained Sequential Linear Quadratic Optimal Control problem. 
The employed algorithm features high efficiency through a continuous-time formulation that benefits from adaptive step-size integrators and through linear complexity in the number of integration steps. 
In a first application example, we solve kinematic trajectory planning problems for a 26~DoF wheeled robot. In a second example, we apply Constrained SLQ to a real-world mobile manipulator in a receding-horizon optimal control fashion, where we obtain optimal controllers and plans at rates up to 100~Hz.
\end{abstract}
%

\section{Introduction}
\label{sec:Introduction}
\subsection{Motivation}
\label{sec:motivation}
Many robotic systems feature inherent motion constraints, such as wheels or tracks, which result in non-holonomic constraints. Non-holonomic constraints are non-integrable and of the form $\vg(\vq, \dot \vq, t) = 0$, where $\vq$ denotes the generalized coordinates of the system. 
Informally speaking, a non-holonomic constraint restricts instantaneous motion in certain directions and therefore leads to limited maneuverability.

\begin{figure}
\subfloat[The In Situ Fabricator 1]{\includegraphics[width = 0.265\textwidth]{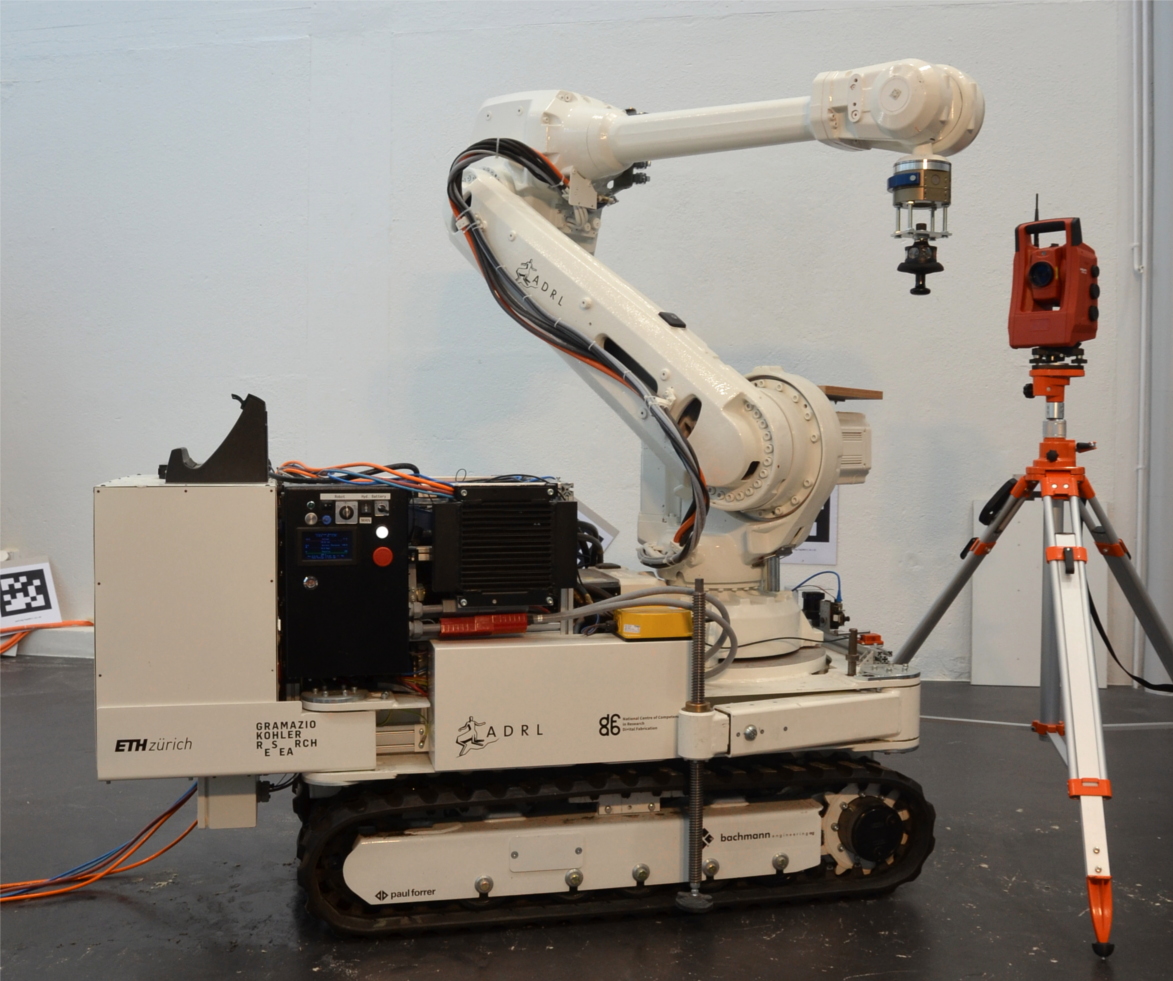} \label{fig:experimental_setup}}
\subfloat[The In Situ Fabricator 2]{\includegraphics[width = 0.212\textwidth]{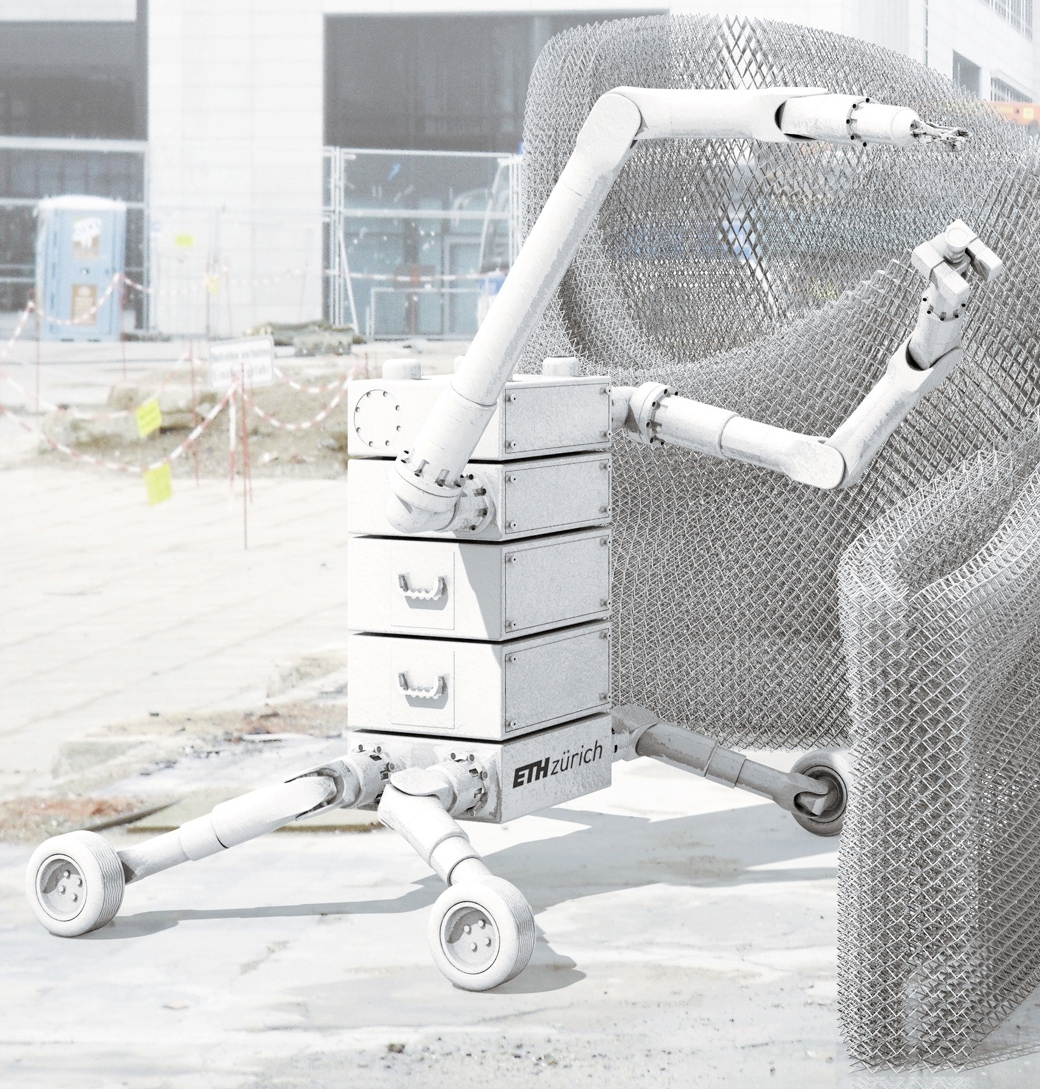}\label{fig:if2}}
\caption{
(a) The In Situ Fabricator 1 is a 1.5 ton mobile manipulator designed for building construction and digital fabrication. It is equipped with a standard industrial robot arm and a hydraulically driven base. \newline
(b) Conceptual rendering of the In Situ Fabricator 2, which is currently under development. 
}\label{fig:in_situ_fabricators}
\end{figure}

Simple non-holonomic systems, such as planar tracked robots and car-like robots with a small number of Degrees of Freedom (DoF) may be intuitive to steer. However, for complex robots like the one shown in Fig.~\ref{fig:if2}, which features a multitude of non-holonomic constraints, we need fast computational methods to generate trajectories.
Fig.~\ref{fig:if2} shows a concept for the ``In Situ Fabricator 2" (IF2), a robot for digital fabrication and on-site building construction, which is currently under development. Because of the broad spectrum of requirements and the need for superior maneuverability, its base is equipped with four legs - each with several degrees of freedom (translational and rotary joints) plus wheels. The depicted robot's base, for instance, features 16~rotational and 4~translational joints.
One of the challenges that we are facing in the development of this machine, is how to efficiently design trajectories and feedback controllers that are consistent with the non-holonomic constraints.
Another requirement in mobile manipulation is the ability to reposition a robot while its end effector remains at a desired pose or follows a desired motion, which falls in the category of operational-space tracking~\cite{khatib1987unified}. 
In this paper, we introduce the Constrained Sequential Linear Quadratic Optimal Control algorithm (Constrained SLQ) to the field of kinematic trajectory planning.  This allows to efficiently solve non-holonomic planning problems subject to operational-space tracking constraints with linear time complexity and provides time-varying kinematic feedback laws.

Although we neglect a consideration of the dynamics, the problem posed in this paper is still relevant to a large number of robotic systems available today: the majority of the currently commercially available (mobile) industrial manipulators are position/velocity-controlled. Furthermore, any robot with joint position sensing can in principle be used in a position/velocity controlled mode.

This paper is structured as follows. In the remainder of Section~\ref{sec:Introduction}, we discuss related work and state the contributions of this paper. In Section~\ref{sec:constrained_slq}, we introduce our version of the Constrained SLQ Algorithm, introduce the kinematic problem formulation and explain our receding horizon optimal control setup. We present an example of trajectory planning for a complex wheeled robot in Section~\ref{sec:Planning_IF2} and apply Constrained SLQ in receding horizon optimal control fashion to a real tracked mobile manipulator in Section~\ref{sec:IF1_MPC}.
In Section~\ref{sec:Discussion}, we summarize and discuss the results. An outlook on future work concludes this paper in Section~\ref{sec:Outlook}.

\subsection{State of the art and related work}
\label{sec:related_work}
Kinematic planning in its different flavours has been treated extensively and in great detail in the literature. For a broad, general overview about planning algorithms, refer to~\cite{lavalle2006planning}. Overviews particularly treating non-holonomic planning can be found in~\cite{laumond1998guidelines, jean2014nonholonomic}.
Well-established methods for kinematic planning of non-holonomic systems include the different variants of sampling-based algorithms like Rapidly-Exploring Random Trees (RRT) or Probabilistic Road Map (PRM) methods, Graph Search methods or Nonlinear Programming (NLP) approaches, each of them with specific strengths and drawbacks.
NLP approaches often scale unfavourably with the problem time horizon. SQP in its basic form, for instance, scales with $O(n^3)$. 
Standard RRTs and PRMs often perform fast, $O(n\text{log}(n))$, but may require post-processing steps like smoothing. Also, they may produce solutions that are kinematically feasible but not guaranteed to be optimal.
Importantly, in~\cite{shkolnik2010sample} it is argued that many sample-based planners rely on distance metrics which might not work well when a system has differential constraints -- which can become a problematic point for robots with many non-holonomic constraints, such as the IF2.
In~\cite{pivtoraiko2009differentially}, an improved differentially constrained search space is constructed, which resolves that problem. 
A systematic analysis of the optimality and complexity parameters for sampling-based path planning algorithms is given in~\cite{karaman2011sampling}. Note that extended algorithms like RRT* and PRM* are asymptotically optimal and computationally efficient. For example, RRT* shows $O(n)$ query time complexity, however still has processing time complexity $O(n\text{log}(n))$.
While sampling-based planners have gained great popularity, other approaches have also been applied successfully to non-holonomic systems. For example, in~\cite{rollingJustin} and~\cite{dietrich2011singularity}, the essentials of kinematic control and planning for the DLR's ``Rollin' Justin" wheeled robot are presented, which is based on feedback linearization. 

Note that in this paper, we plan in the space of generalized coordinates -- however the reference trajectories to be tracked can be given in operational space.
Relevant works for operational space tracking for non-holonomic robots include~\cite{oriolo2005motionplanning}, where motions are computed along a given end effector path subject to non-holonomic base constraints using a sampling-based planner. Other methods for the planning of operational space tracking tasks were presented in~\cite{bianco2014real} and~\cite{huang2006tracking}.

A connection between optimal control and inverse kinematics has been made in~\cite{geoffroy2014inverse}, where it is argued that inverse kinematics can be seen as a special case of an optimal control formulation, when the preview horizon collapses. 
Our work extends these results such that we can now use optimal control in scenarios with a non-zero preview horizon, where one would traditionally still revert to using inverse kinematics. 

The idea of equality-constrained SLQ has originally been introduced in~\cite{sideris10}, using a formulation in discrete-time. We present a derivation of the algorithm in continuous time in~\cite{farshidian16efficient}, which is conceptually different. The continuous time version allows us to use adaptive step-size integrators, which helps to achieve shorter run-times in practice.

\subsection{Contributions of this paper}
\label{sec:contributions_of_this_paper}
The aim of this work is not to replace existing kinematic planners in large planning problems with possibly cluttered environments, since our approach is based on a convex optimization framework. 
To some degree, it can handle non-convex solution sets, however in cluttered environments, the problem is ill-posed and other types of planners are more suitable.
In a local regime however, our approach has favourable complexity properties, i.e. linear time complexity~$O(n)$. Additionally, it computes kinematic feedback laws that are compliant with the constraints, and the algorithm does not suffer from the shortcomings that many algorithms which employ inverse kinematics exhibit, i.e. when approaching singularities.
Therefore, this paper complements existing work on kinematic planning and control of non-holonomic mobile manipulators by introducing Constrained SLQ to the field of kinematic planning. 
We show examples where we use Constrained SLQ to efficiently plan motions for robots with complex kinematics, non-holonomic constraints and given end effector trajectories. Furthermore, we show the performance of the algorithm in a receding-horizon optimal control experiment on a real tracked mobile manipulator.

\section{Constrained SLQ}
\label{sec:constrained_slq}
\subsection{Algorithmic overview}
Constrained SLQ is based on Dynamic Programming (DP), which designs both a feedforward plan and a feedback controller. 
While conventional DP methods are effective tools for solving optimal control problems, they do not scale favorably with the time horizon. 
However, a class of DP algorithms known as SLQ algorithms exists that scales linearly with the optimization time horizon. 
Our continuous--time SLQ algorithm can handle state and input constraints while the complexity remains~$O(n)$.

In general, NLP--based planning algorithms require the discretization of the infinite dimensional, continuous optimization problem to a finite dimension NLP. This discretization is often carried out using heuristics, which can result in numerically poor or practically infeasible solutions. Our algorithm, by contrast, is a continuous--time method which uses variable step-size ODE solvers in its forward and backward passes. Given the desired accuracy, it can automatically discretize the problem using the error control mechanism of the variable step-size ODE solver. 
Informally speaking, this allows the solver to indirectly control the distance between the ``nodes" in the feedforward and feedback trajectory. In practice, this decreases the runtime of an iteration, since the number of calculations decreases.  

Another aspect of our algorithm is that it produces feedback plans. While a feedforward plan provides a single optimal open-loop trajectory, a feedback scheme generalizes the plan to the vicinity of the current solution.
Our algorithm uses linear feedback controllers for this purpose, hence we obtain time-varying control laws of form
\begin{equation}
\vu(\vx,t) = \vu_{ff}(t)+\vK(t)\vx(t) \ \text{.}
\label{eq:control_law_slq}
\end{equation}

We formulate the constrained optimal control problem as

\begin{small}
\begin{align}
    & \min_{\vu (\cdot)} \left\{ \Phi(\vx (t_f)) + \int_{t_0}^{t_f} L(\vx, \vu, t) dt \right\} \notag
    \\
    & \textrm{subject to}  \notag
    \\
    & \dot{\vx} =  \vf(\vx, \vu) \hspace{11mm} \vx(t_0) = \vx_0  \notag   
    \\
    & \vg_1(\vx, \vu, t) = 0 \qquad  \vg_2(\vx, t) = 0                            
    \label{eq:general_op}
\end{align}
\end{small}%
The nonlinear cost function consists of a terminal cost $\Phi$ and an intermediate cost $L$. $\vf(\cdot)$ is the system differential equation. $\vg_1(\cdot)$ and $\vg_2(\cdot)$ are the state-input and pure state constraints, respectively. 

The constrained SLQ algorithm is an iterative method which approximates the nonlinear optimal control problem with a local Linear Quadratic (LQ) subproblem in each iteration, and then solves it through a Riccati-based approach. 

The first step of each iteration is a forward integration of the system dynamics using the last approximation of the optimal controller. Note that in the very first iteration, the algorithm needs to be initialized with a stable control policy. 
Next, it calculates a quadratic approximation of the cost function over the nominal state and input trajectories obtained from the forward integration. 
The cost's quadratic approximation is

\begin{small}
\begin{align}
& \widetilde{J} = \widetilde{\Phi}(\vx({t_f}))+ \int_{t_0}^{t_f} { \widetilde{L}(\vx,\vu,t)dt} \notag \\
&\widetilde{\Phi}(\vx(t_f)) = \ q_{t_f} + \vq_{t_f}^\top \delta\vx + \frac{1}{2} \delta\vx^\top \vQ_{t_f} \delta\vx \notag \\
& \widetilde{L}(\vx,\vu,t) = q(t) + \vq(t)^\top \delta\vx + \vr(t)^\top \delta\vu  + \delta\vx^\top \vP(t) \delta\vu  \notag \\
& \hspace{17mm} + \frac{1}{2} \delta\vx^\top \vQ(t) \delta\vx + \frac{1}{2} \delta\vu^\top \vR(t) \delta\vu  
\label{eq:cost_quadratic_approximation}
\end{align} 
\end{small}%
where $q(t)$, $\vq(t)$, $\vr(t)$, $\vP(t)$, $\vQ(t)$, and $\vR(t)$ are the coefficients of the Taylor expansion of the cost function in Equation~\eqref{eq:general_op} around the nominal trajectories. $\delta\vx$ and $\delta\vu$ are the deviations of state and input from the nominal trajectories. Constrained SLQ also uses linear approximations of the system dynamics and constraints in Equation~\eqref{eq:general_op} around the nominal trajectories as follows:
\begin{align} 
& \delta\dot{\vx} = \vA(t)\delta\vx + \vB(t)\delta\vu \notag \\
& \vC(t)\delta\vx + \vD(t)\delta\vu + \ve(t) = \mathbf{0} \notag \\
& \vF(t)\delta\vx + \vh(t) = \mathbf{0} 
\label{eq:dynamics_linear_approximation}
\end{align}
Based on this LQ approximation, we use a generalized, constrained LQR algorithm to find an update to the feedback-feedforward controller. Constrained SLQ is described in Algorithm~\ref{alg:cslq}. For a more detailed discussion of the algorithm's derivation, refer to~\cite{farshidian16efficient}.

\begin{algorithm}[htpb] 
\caption{Constrained SLQ Algorithm} 
\label{alg:cslq}
\begin{algorithmic} \scriptsize \STATE \textbf{Given} 
\STATE - The optimization problem in Equation~\eqref{eq:general_op}
\STATE - Initial stable control law, $\mathbf{\vu}(\vx,t)$ 
\REPEAT \STATE - forward integrate the system equations using adaptive step-size integrator: 
\STATE $\hspace{2mm} \tau: \overline{\vx}(t_0),\overline{\vu}(t_0),\overline{\vx}(t_1),\overline{\vu}(t_1)\dots\overline{\vx}(t_{N-1}),\overline{\vu}(t_{N-1}),\overline{\vx}(t_N=t_f)$
\STATE - Quadratize cost function along the trajectory $\tau$
\STATE - Linearize the system dynamics and constraints along the trajectory $\tau$
\STATE - Compute the constrained LQR problem coefficients%
\STATE \qquad $\vD^\dagger = \vR^{-1} \vD^\top(\vD \vR^{-1} \vD^\top)^{-1}, \quad \widetilde\vA = \vA - \vB \vD^\dagger \vC $
\STATE \qquad $\widetilde\vC = \vD^\dagger \vC, \quad \widetilde \vD = \vD^\dagger \vD, \quad \widetilde \ve = \vD^\dagger \ve $
\STATE \qquad $\widetilde \vQ = \vQ + \widetilde\vC^\top \vR \; \widetilde\vC  - \vP \widetilde\vC - (\vP \widetilde\vC)^\top + \vF^\top \vF $
\STATE \qquad $ \widetilde\vq = \vq - \widetilde \vC^\top \vr + \vF^\top \vh, \quad \widetilde \vR = (\vI - \widetilde \vD)^\top \vR (\vI - \widetilde \vD) $
\STATE \qquad $\widetilde\vL = \vR^{-1} ( \vP^\top + \vB^\top \vS ) $
\STATE \qquad $\widetilde\vl = \vR^{-1} ( \vr + \vB^\top \vs ), \quad \widetilde \vl_e = \vR^{-1} \vB^\top \vs_e $
\STATE - Solve the final--value Riccati-like equations
\STATE \qquad $- \dot \vS = \widetilde\vA^\top \vS + \vS^\top \widetilde\vA - \widetilde\vL^\top \widetilde\vR \; \widetilde\vL + \widetilde\vQ  \hspace{12mm} \vS(t_f)=\vQ_f  \hspace{5mm}$
\STATE \qquad $- \dot \vs = \widetilde\vA^\top \vs - \widetilde\vL^\top \widetilde\vR \; \widetilde\vl + \widetilde\vq  \hspace{25mm} \vs(t_f)=\vq_f$
\STATE \qquad $- \dot \vs_e = \widetilde \vA ^\top \vs_e - \widetilde\vL^\top \widetilde\vR  \; \widetilde\vl_e + ( \widetilde\vC - \widetilde\vL )^\top \vR \; \widetilde\ve \hspace{4.5mm} \vs_e(t_f)=\boldmath{0}$
\STATE \qquad $- \dot{s} = q - \widetilde\vl^\top \widetilde\vR \; \widetilde\vl, \hspace{34mm} s(t_f)=q_f$
\STATE - Compute the controller update
\STATE \qquad $ \vL = -(\vI - \widetilde\vD) \widetilde\vL - \widetilde\vC \hspace{45mm}$
\STATE \qquad $ \vl = -(\vI - \widetilde\vD) \widetilde\vl$
\STATE \qquad $ \vl_e =  -(\vI - \widetilde\vD) \widetilde\vl_e - \widetilde\ve$
\STATE \qquad $ \delta\vu = \alpha \vl + \alpha_e \vl_e +  \vL \vx$
\STATE - Optimize $\alpha_e$ and $\alpha$ using a line search scheme
\STATE - Update the controller
\STATE \qquad $ \vK \leftarrow \vL 
\hspace{45mm}$
\STATE \qquad $ \vu_{ff} \leftarrow \overline{\vu} + \alpha \vl + \alpha_{e} \vl_e - \vL \overline{\vx}
\hspace{35mm}$
\UNTIL{convergence or maximum number of iterations}
\end{algorithmic} 
\end{algorithm}

\subsection{Formulation of the kinematic planning and control problem}
\label{sec:formulation_kin_problem}
For the kinematic planning and control problems considered in this paper, we define the state $\vx$ as the robot's generalized coordinates $\vq$, and the input $\vu$ as their time derivatives (velocities). We assume that the system's generalized coordinates are fully observable at any time.

The initial control law supplied to the Constrained SLQ algorithm needs to stabilize the system. In the case of kinematic planning, we can obtain this through a constant, zero control input.
For all presented applications, the cost functions are of the form 
\begin{equation}
    \int_{t_0}^{t_f} \vu(t)^\top \vR \vu(t) dt+(\vx(t_f)-\vx_r)^\top\vQ_f(\vx(t_f)-\vx_r) 
    \label{eq:used_cost_function}
\end{equation}
hence the intermediate states are never penalized, only the control inputs and the deviation from the desired terminal state~$\vx_{r}$. The state-input constraints take the form $\vg_1(\vq, \dot \vq, t) = 0$, which covers all non-holonomic constraints, and the state-only constraints take the form $\vg_2(\vq,t) = 0$, which covers holonomic operational-space tracking constraints.

\subsection{Receding horizon optimal control setup}
\label{sec:receding_horizon_opt}
Later in this work, we use Constrained SLQ in a model predictive control fashion. For that purpose, we split the control framework into an outer and an inner loop.

In the outer loop, we iteratively solve the optimization problem~\eqref{eq:general_op} with a constant time horizon using Constrained SLQ and the current estimate of the system state.
High update rates can be achieved through 
\begin{itemize}
\item ``warmstarting" the algorithm with previous solutions
\item adaption of the integrator tolerances for the forward and backward pass integrations. 
\end{itemize}

The inner loop implements a closed-loop controller~\eqref{eq:control_law_slq} which applies the optimal feedforward and feedback trajectories designed by the outer loop.
In our setup, the inner loop runs at higher frequency than the outer loop and ensures stable feedback while new optimal control trajectories are designed.
For subsampling feedforward and feedback matrices between the plan's nodes, we apply linear interpolation.

This setup provides the advantage that small disturbances of higher frequency can be handled directly by the feedback controller in the inner loop, while the planner in the outer loop reacts to perturbations of larger time-scale and magnitude through replanning.
The presented scheme of receding horizon optimal control has previously been applied to a hexrotor system in~\cite{neunert16hexrotor}, using the \emph{unconstrained} version of SLQ.

\section{Planning for a legged/wheeled Mobile Robot}
\label{sec:Planning_IF2}
\subsection{System and constraint modelling}
We have designed a kinematic model of the legged/wheeled base of the In Situ Fabricator~2, featuring a total of 26~DoF.
Several assumptions are made to simplify the formulation of the kinematic constraints. A generalization is possible but beyond the scope of this paper.
Fig.~\ref{fig:wheel_sketch} shows a sketch of one of the wheels connected to the robot's base via a serial chain of joints and links with joint positions~$\vph$ and velocities~$\dot \vph$.
We assume flat ground and model the wheels as perfect, `infinitely thin' discs, where a unique ground contact point $P$ exists. Therefore, the distance between $P$ and the  wheel center point $C$ is constant and identical to the wheel's radius $r$. The wheel rotates about the z-axis of the last joint's coordinate system $fr_J$ with angular velocity $\vom_{w}^{fr_J}$. We denote the center of the base coordinate system $B$, the world and the base frame are $fr_W$ and $fr_B$, respectively.

We define the state $\vx$ in terms of the robot's generalized coordinates: as the robot's base orientation $\vth_B^{fr_W}$ and position $\vr_B^{fr_W}$ in the world frame, as well as all leg- and wheel joint angles $\vph$.
The control inputs $\vu$ are defined as the local angular- and translational velocities of the trunk $\vom_B^{fr_B}$, $\vv_B^{fr_B}$ and all joint velocities, $\dot \vph_{leg_i}$.
In the following, we do not detail on how to calculate all of the required kinematic quantities, but we point out that many of them can be automatically generated through the robotics code generator~\cite{frigerioCodeGen}, which we used in this work.
Note that the procedure holds for an arbitrarily complex serial chain of links and joints.

\subsubsection{Ground contact point velocities}
The angular and translational velocities of $C$ in the base frame can be obtained via the last joint frame's Jacobian w.r.t. the base, therefore
\begin{equation}
\left[ \vom_C^{{fr_B} \top} \quad  \vv_{C}^{{fr_B} \top} \right] ^\top  = \vJ_C^{fr_B}(\vph)  \cdot \dot \vph \ \text{.}
\label{eq:jacobian_ee_base}
\end{equation}
Considering the geometry of the setup, it is simple to calculate the position vector from $C$ to $P$ in the  wheel joint frame,~$\vr_{CP}^{fr_J}$. 
The local angular velocity of the wheel represented in the wheel joint frame is $\vom_{w}^{fr_J} = [0 \ 0  \ \dot q_w]^\top$. 
Therefore, the contact point's velocity in the base frame reads as
\begin{equation}
 \vv_P^{fr_B} =  \vv_C^{fr_B} + \vR_{fr_J}^{fr_B}(\vph) \cdot \left( \vom_{w}^{fr_J} \times  \vr_{CP}^{fr_J} \right) 
 \label{eq:contact_point_velocity_base}
\end{equation}
and its velocity in the world frame $\vv_P^{fr_W}$ follows as 
\begin{equation}
\vv_P^{fr_W} = \vR_{fr_B}^{fr_W}(\vth) \cdot \left(
\vv_B^{fr_B} + \vv_P^{fr_B}  +  \vom_B^{fr_B} \times \vr_{BP}^{fr_B}
\right)  \  \text{.}
\label{eq:contact_point_velocity_world}
\end{equation}
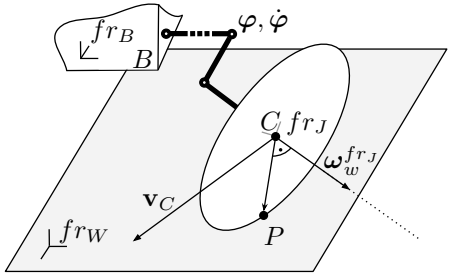
\begin{figure} [tbp]
\centering
\begin{tikzpicture}[scale=1]
\node (pic) at (0,0) {\includegraphics[width = 0.336\textwidth]{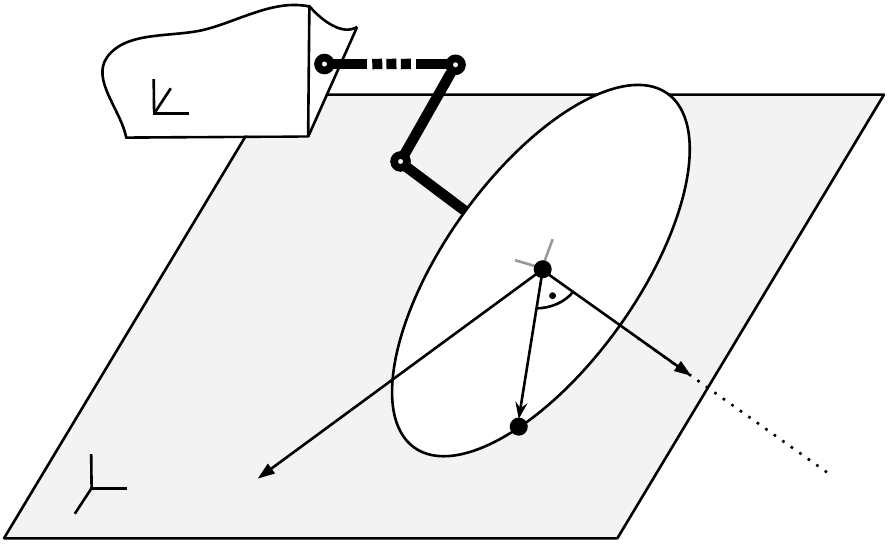} };
\node [right] at (-1.96, 1.47){$fr_B$};
\node [right] at (-1.4, 1.1){$B$};
\node [right] at (-2.38, -1.19){$fr_W$};
\node [right] at (0.65, 0.25){$fr_J$};
\node [right] at (1.15, -0.25){$\vom_w^{fr_J}$};
\node [right] at (0.35, -1.26){$P$};
\node [right] at (0.3, 0.28){$C$};
\node [right] at (-1.26, -0.8){$\vv_C$};
\node [right] at (0.0, 1.61){$\vph, \dot \vph$};
\end{tikzpicture}
\caption{Sketch of an ideal wheel attached to the robot's trunk through an arbitrary serial chain of joints and links.}
\label{fig:wheel_sketch}
\end{figure}

\subsubsection{Wheel constraints}
A rigid body which rolls without sliding fulfils the `rolling condition', which links its angular velocity with the translational velocity of its center of rotation. The rolling condition requires the instantaneous velocity of the ground contact point $P$ in the contact plane to be zero. 

Since the wheel is not supposed to lift off of the ground, we desire the contact point's velocity in world z-direction to be zero, too. This results in a combined, straightforward formulation for the non-holonomic and ground contact constraint for the wheel, which we can directly write as a state-input constraint for the problem introduced in~\eqref{eq:general_op}:
\begin{equation}
\vg_1(\vq, \dot \vq, t) \ := \ \vv_P^{fr_W} \ = \ 0
\label{eq:final_nh_constraint_if2}
\end{equation}
For each leg, this constraint has dimension~3. For the four-legged system, the vector of wheel constraints is therefore of dimension~12. 
In our C++ implementation, we compute the constraint derivatives $\nicefrac{\partial \vg_1 }{\partial \vx}$ and $\nicefrac{\partial \vg_1 }{\partial \vu}$ through automatic differentiation with \mbox{CppAD}~\cite{bell2012cppad,Giftthaler2017autodiff}.

\begin{figure*}[tb]
\setlength{\tabcolsep}{-1pt}    
\renewcommand{\arraystretch}{0.1}
\begin{tabular}{c}  
\subfloat[Task IF2-A: the robot repositioning diagonally, seen from top. The trunk goes from $\left( x,y \right) = \left( 0,0 \right)$ to $\left(1,1\right)$.]{
\begin{tabular}{ccccc}  
    \includegraphics[width = 0.196\textwidth]{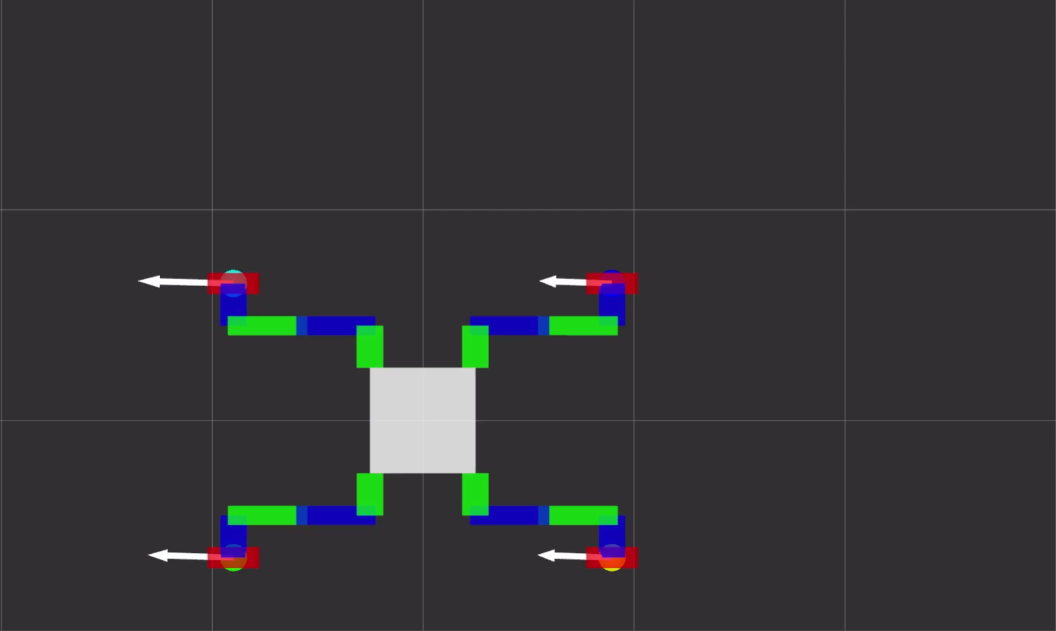} \hfill
&   \includegraphics[width = 0.196\textwidth]{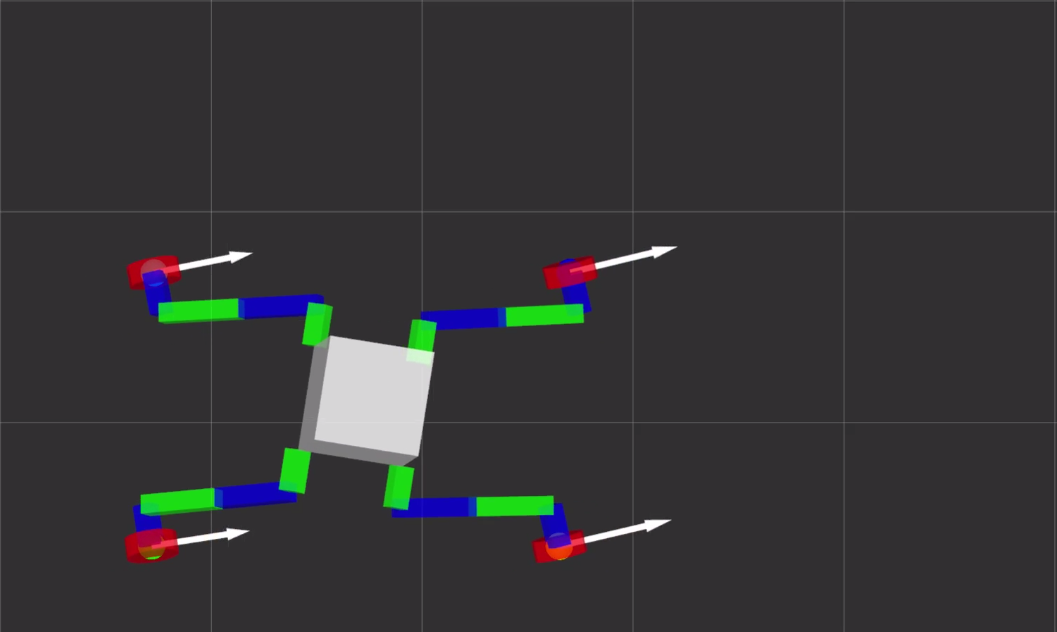} \hfill
&   \includegraphics[width = 0.196\textwidth]{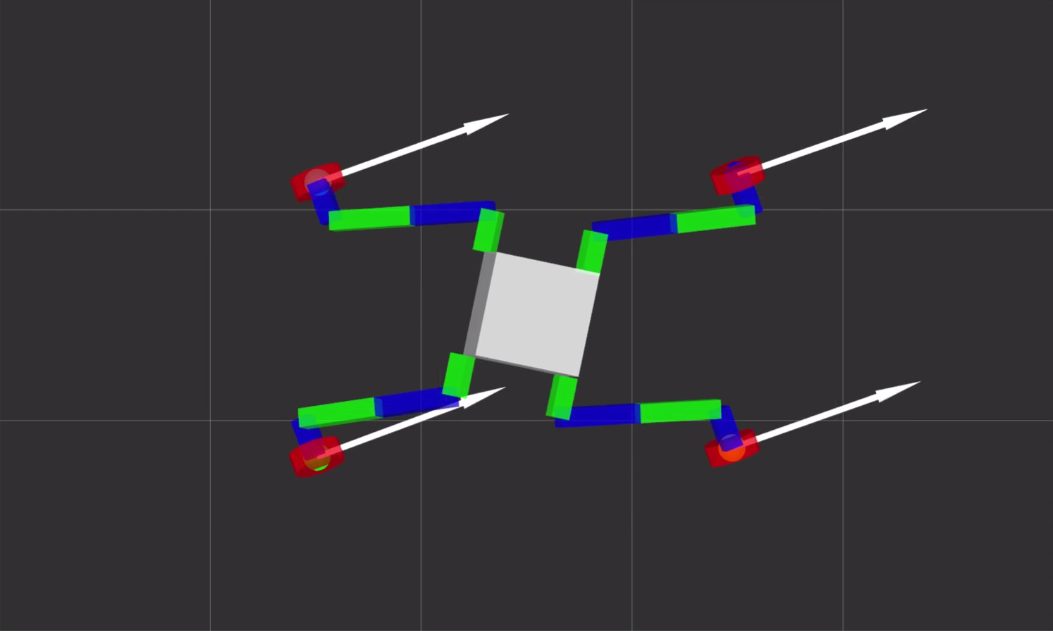} \hfill
&   \includegraphics[width = 0.196\textwidth]{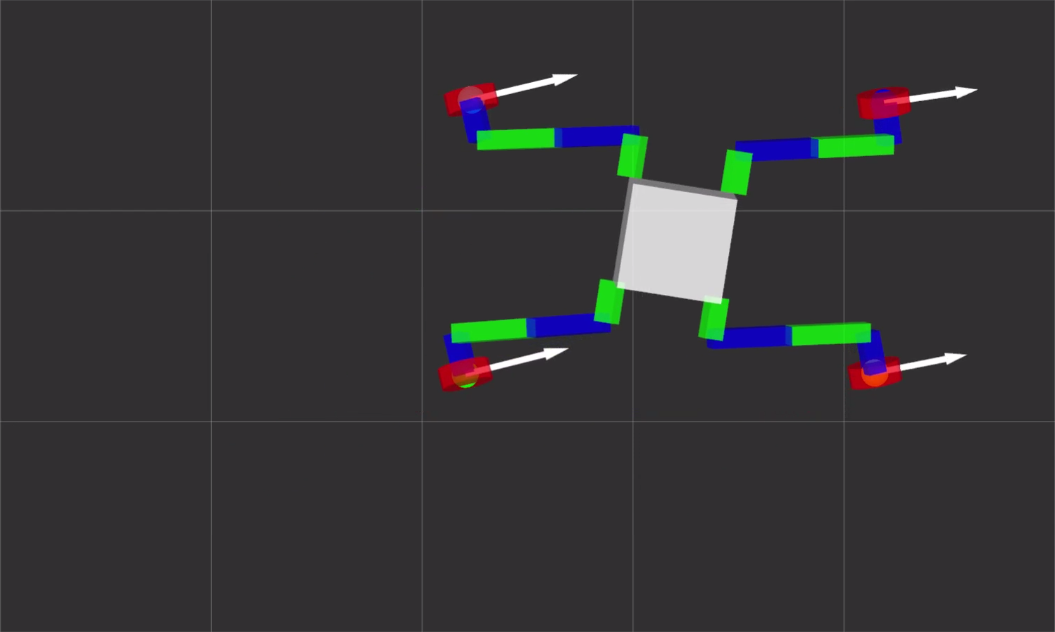} \hfill
&   \includegraphics[width = 0.196\textwidth]{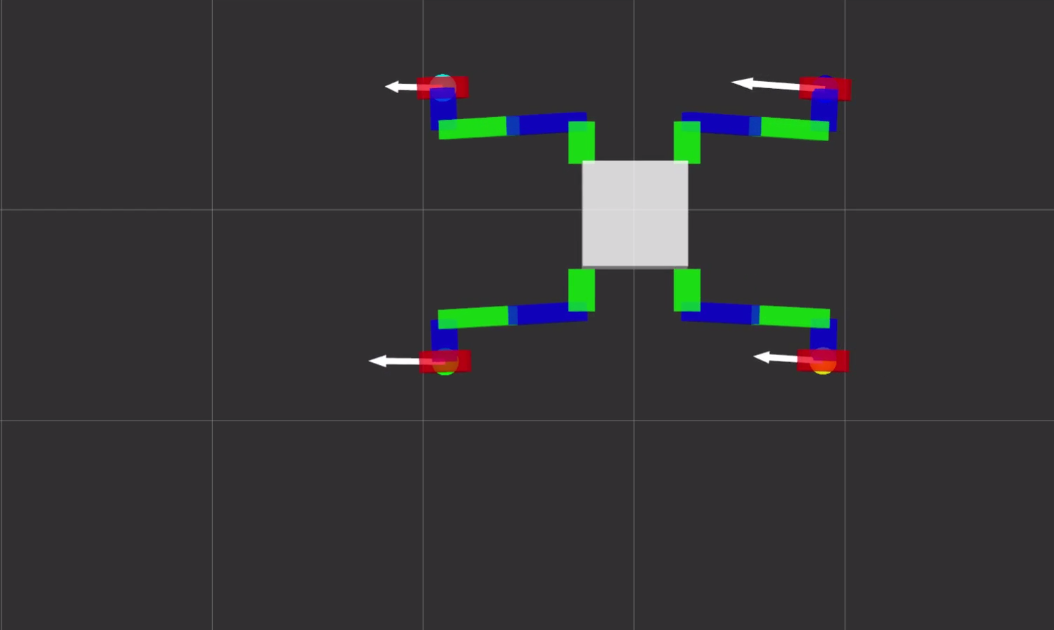} \hfill
\end{tabular}
\label{fig:IF2_repositioning} 
}
\hfill
\vspace*{-0.7em} 
\\
\subfloat[Task IF2-B: the robot performing a manoeuvre where it translates 1.0~meter in $x$ and rotates $180^\circ$, seen from a slanted view from above.]{
\begin{tabular}{ccccc}  
    \includegraphics[width = 0.196\textwidth]{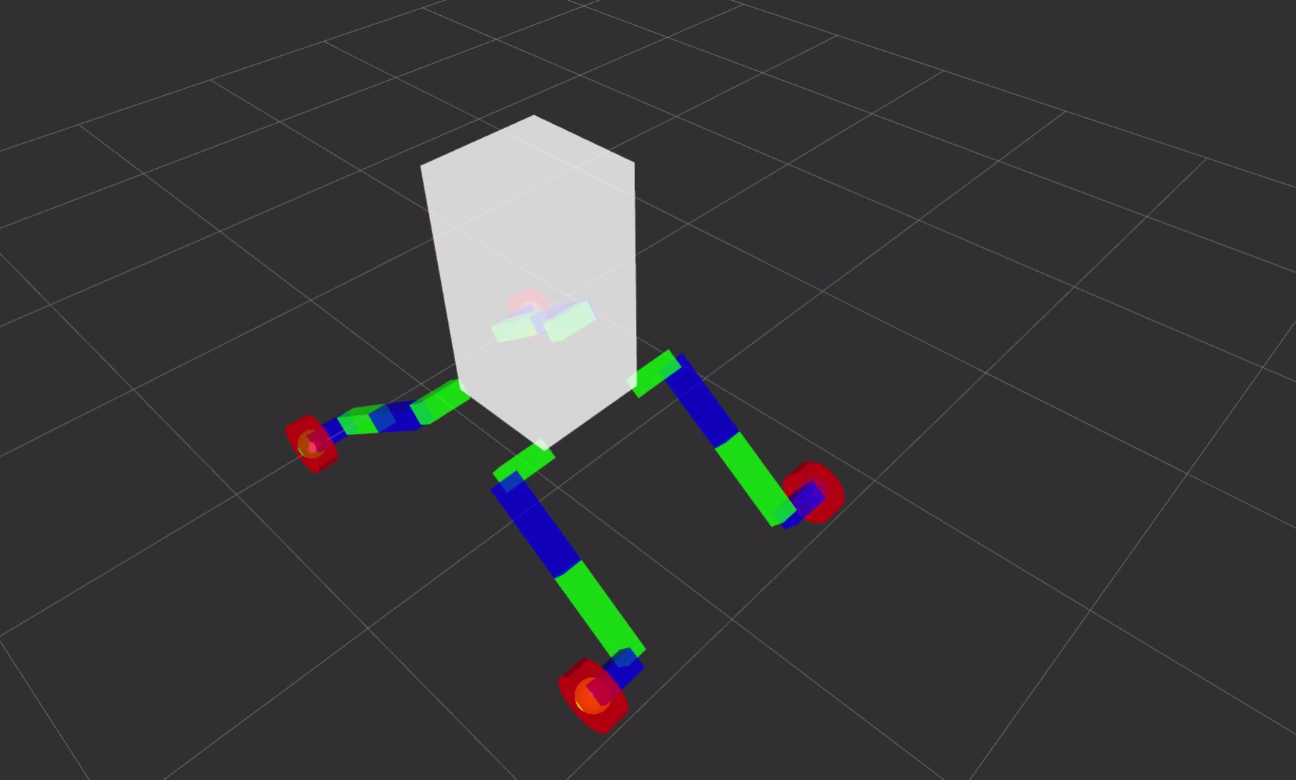} \hfill
&   \includegraphics[width = 0.196\textwidth]{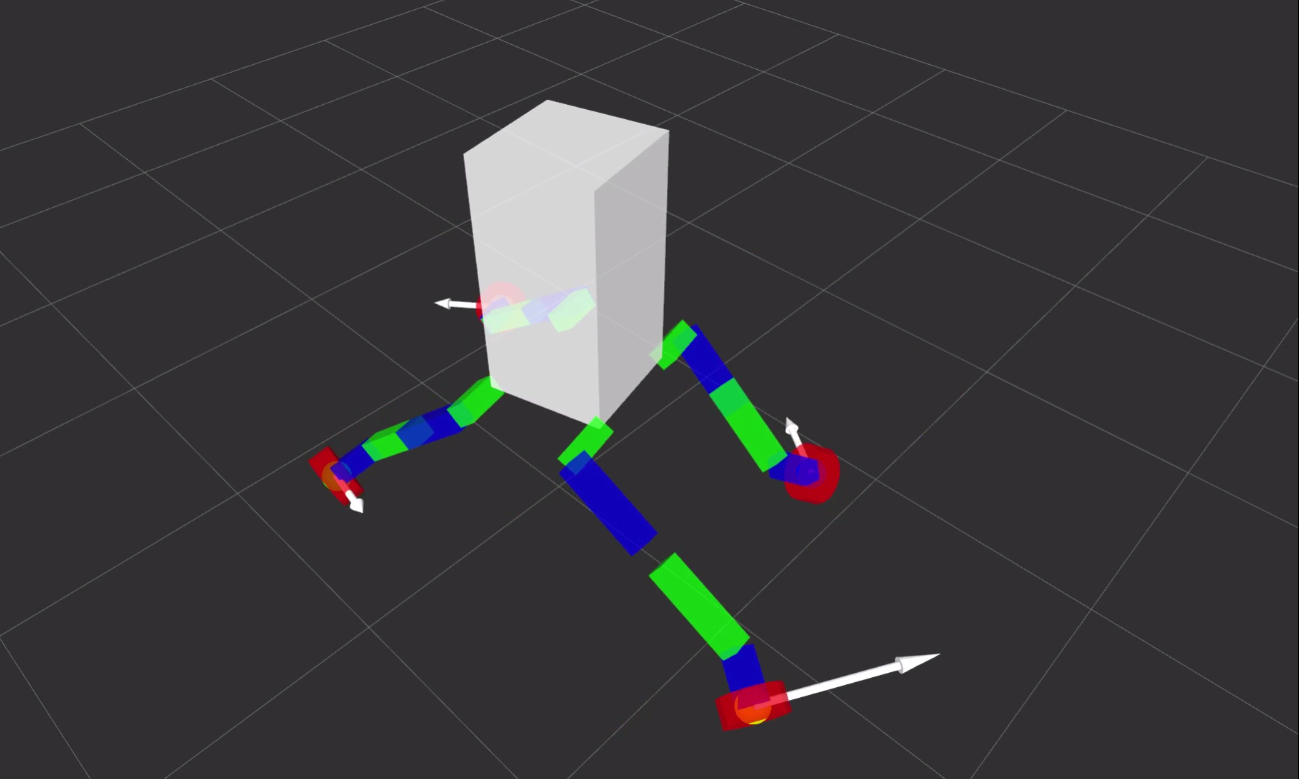} \hfill
&   \includegraphics[width = 0.196\textwidth]{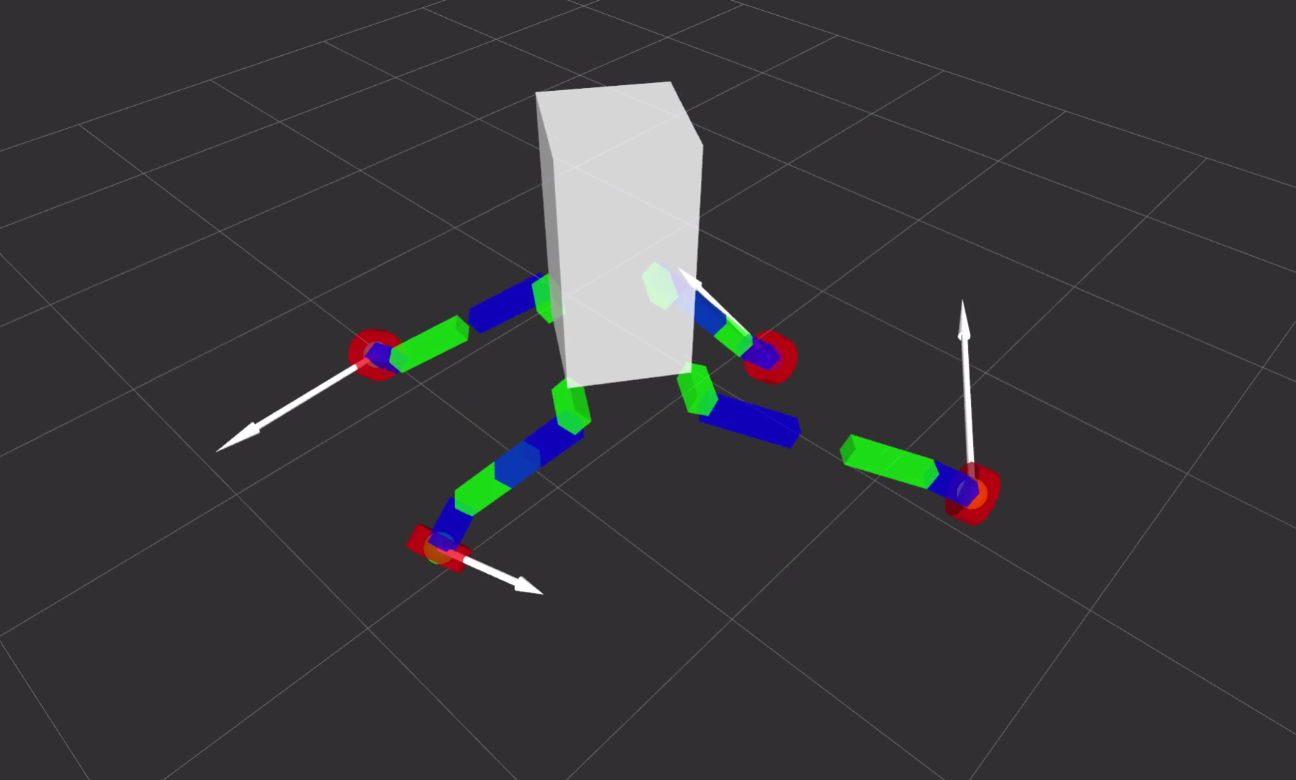} \hfill
&   \includegraphics[width = 0.196\textwidth]{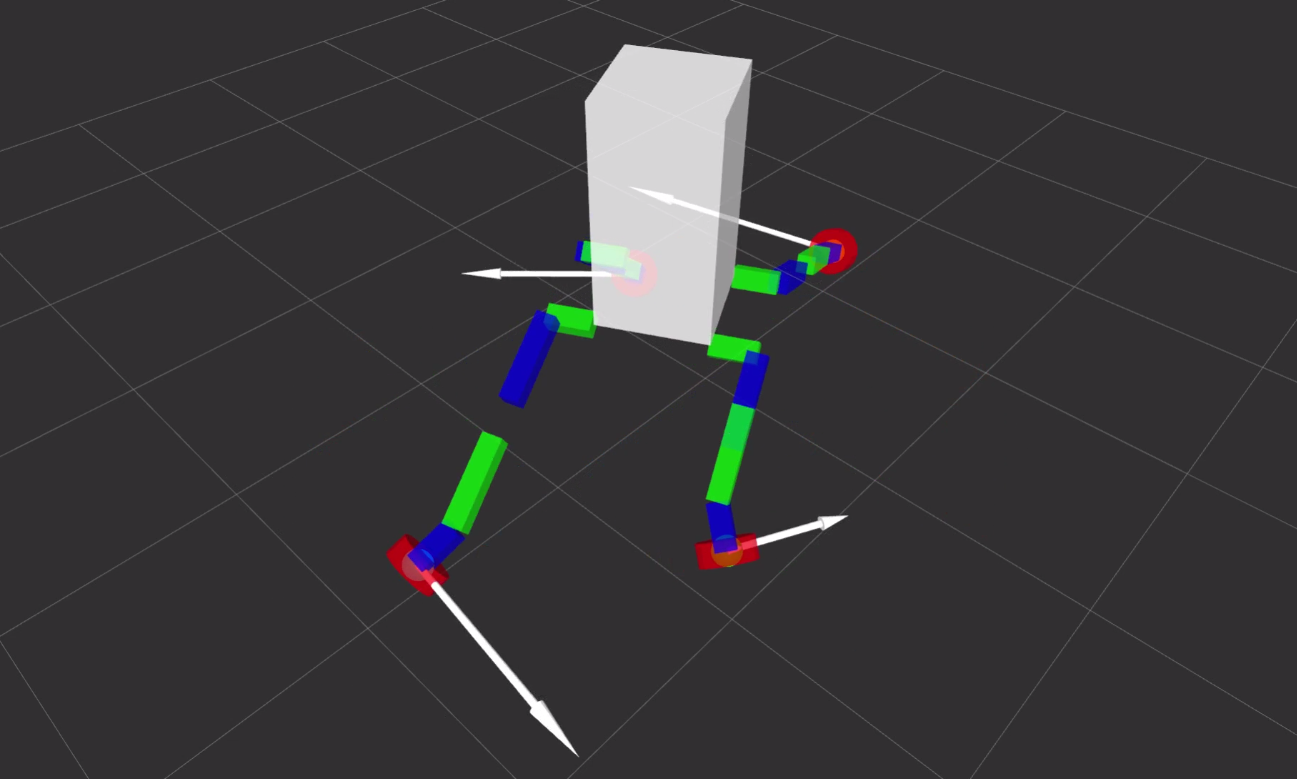} \hfill
&   \includegraphics[width = 0.196\textwidth]{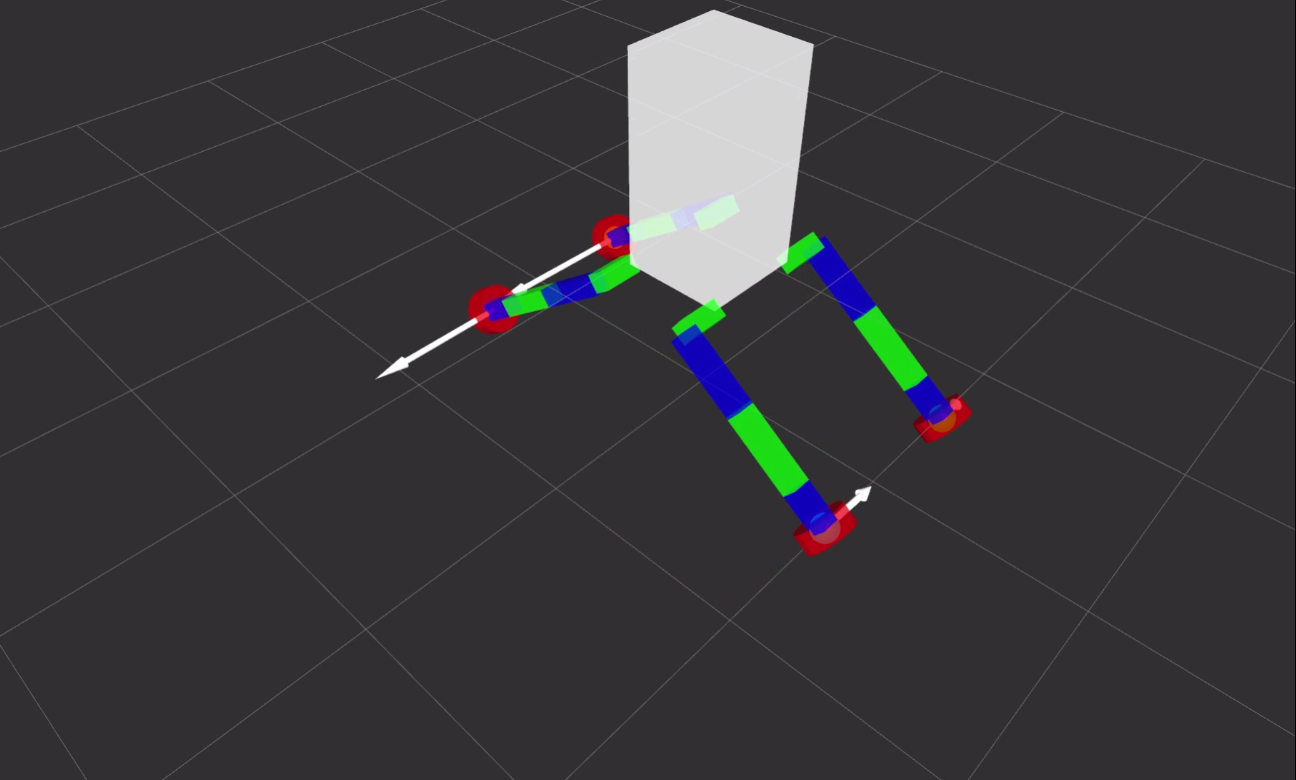} \hfill
\end{tabular}
\label{fig:IF2_repositioning_and_rot} 
}
\hfill
\vspace*{-0.7em} 
\\
\subfloat[Visualization of a whole-body trajectory for the IF1 for an end-effector tracking problem.]{
\begin{tabular}{ccccc}  
    \includegraphics[width = 0.196\textwidth]{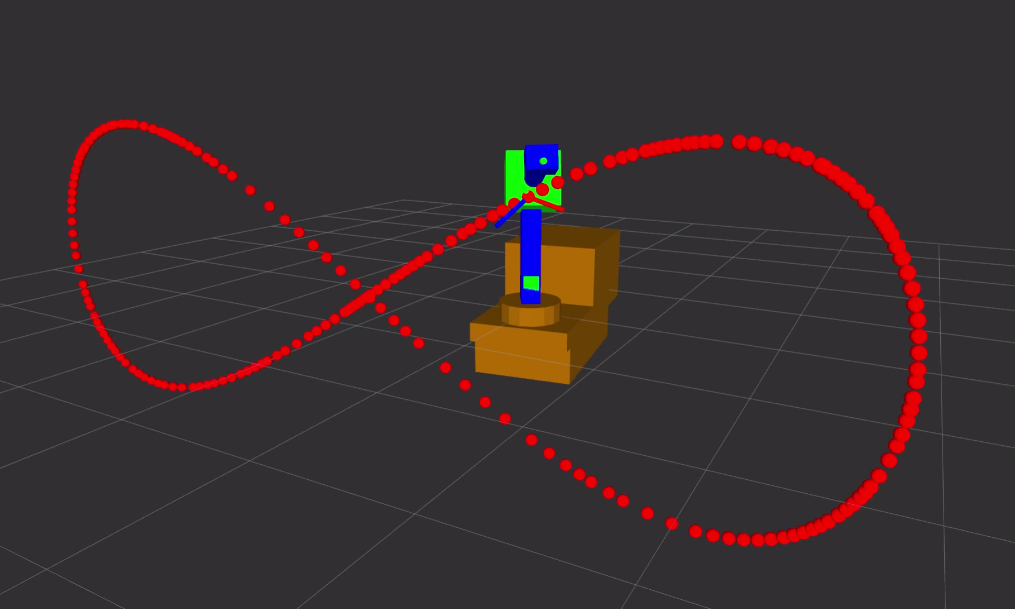} \hfill
&   \includegraphics[width = 0.196\textwidth]{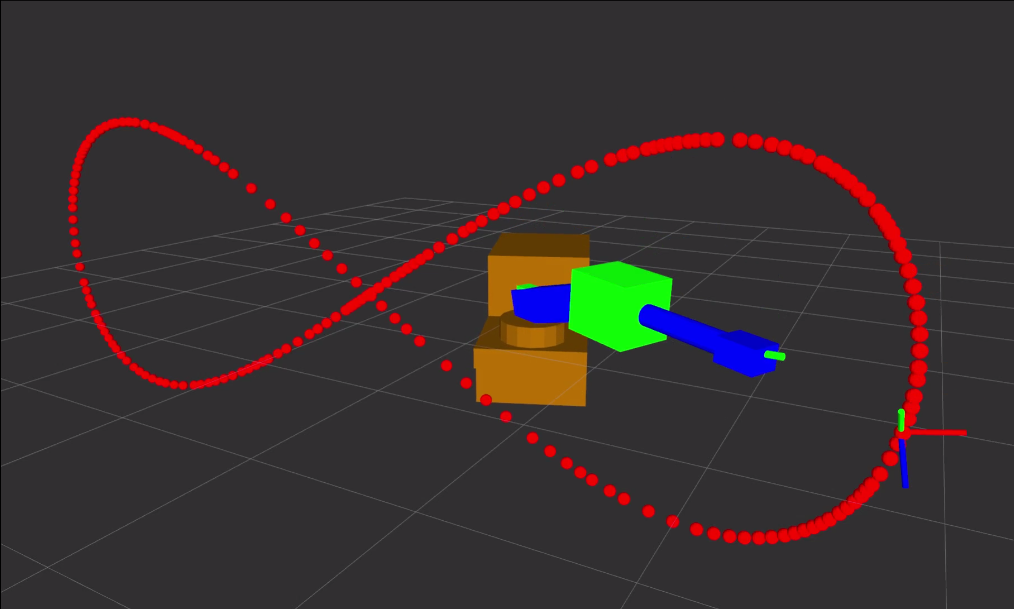} \hfill
&   \includegraphics[width = 0.196\textwidth]{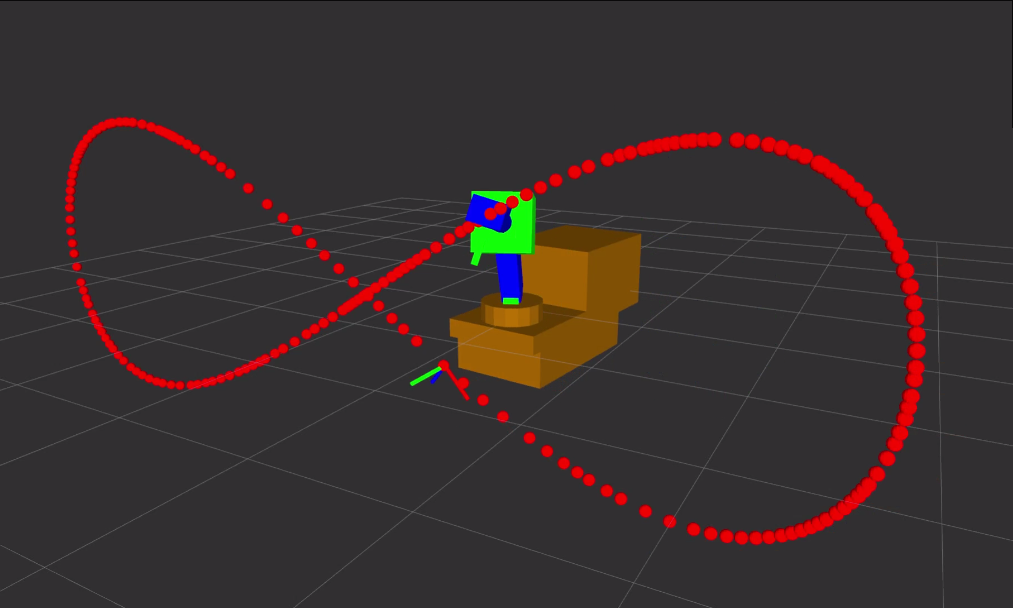} \hfill
&   \includegraphics[width = 0.196\textwidth]{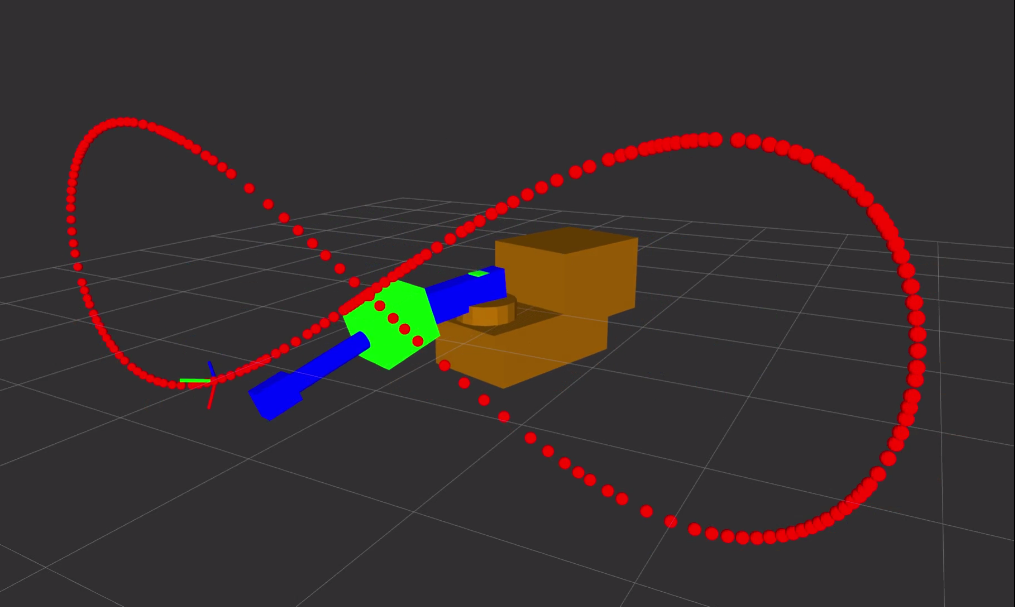} \hfill
&   \includegraphics[width = 0.196\textwidth]{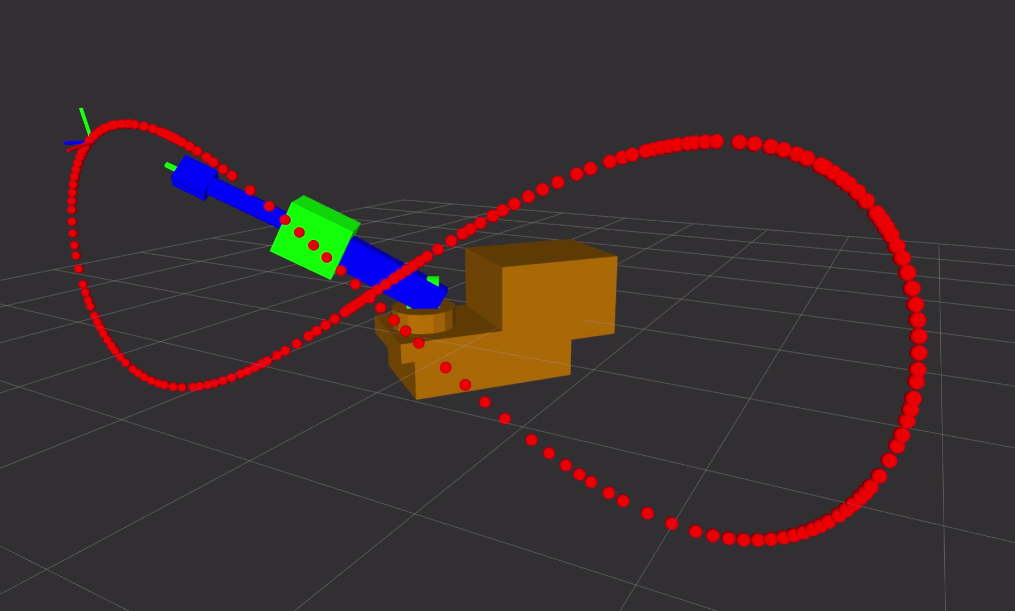} \hfill
\end{tabular}
\label{fig:IF1_figure8} 
}
\hfill 
\vspace*{-0.7em} 
\\
\subfloat[Whole body motion of the real IF1 with end effector constraint. The end effector stays in place while the robots base performs a motion pattern.]{
\begin{tabular}{ccccc}  
    \includegraphics[width = 0.196\textwidth]{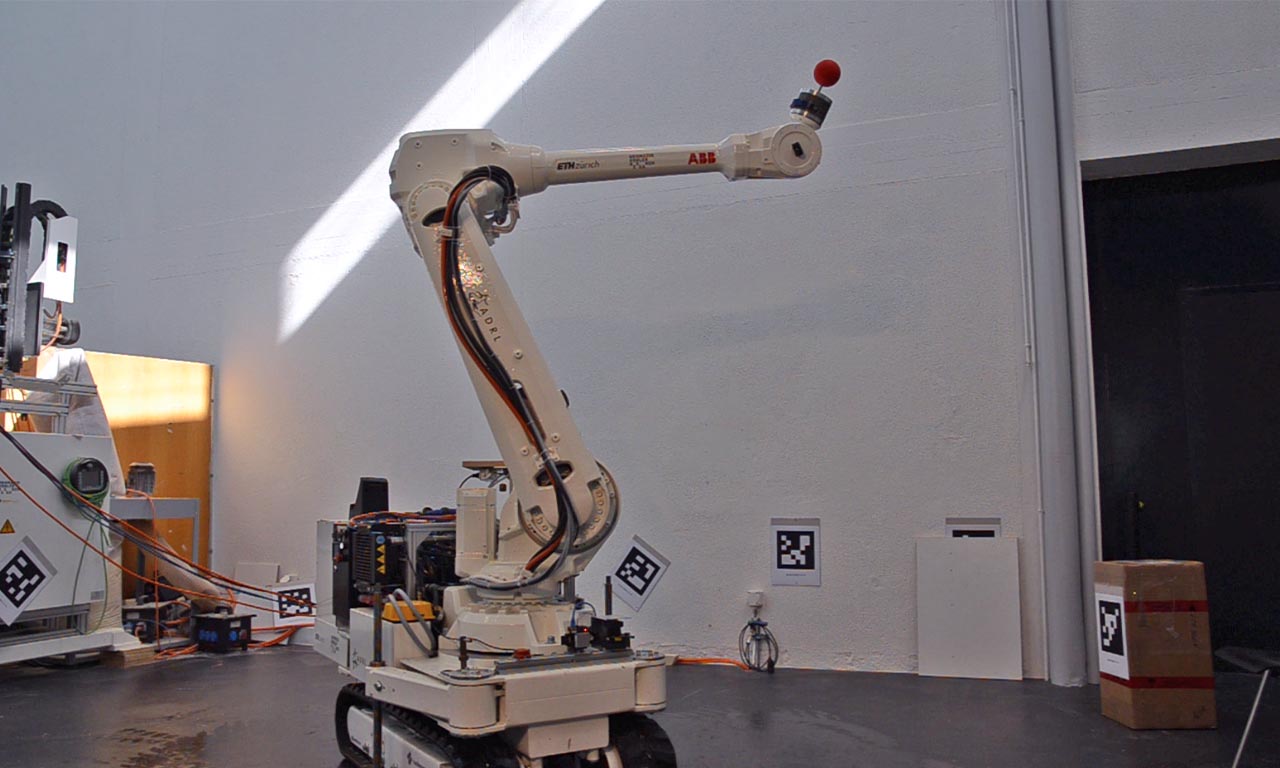} \hfill
&   \includegraphics[width = 0.196\textwidth]{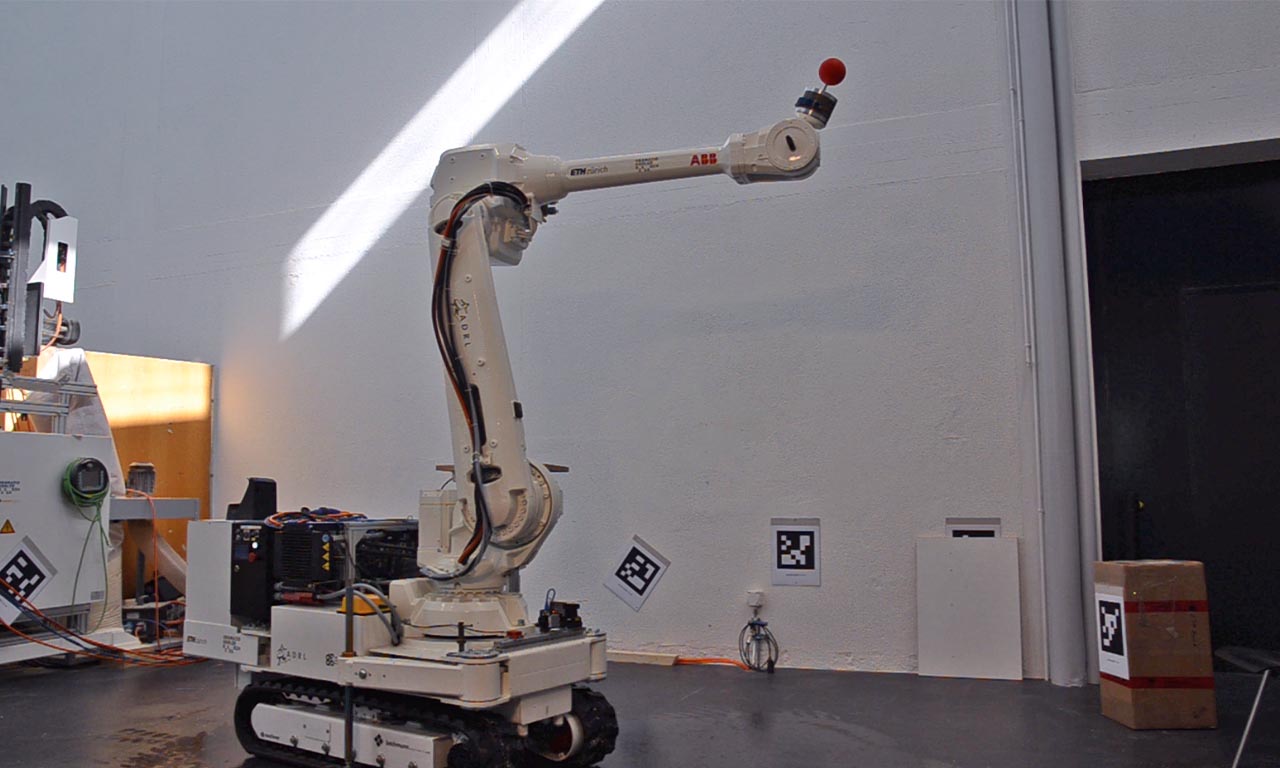} \hfill
&   \includegraphics[width = 0.196\textwidth]{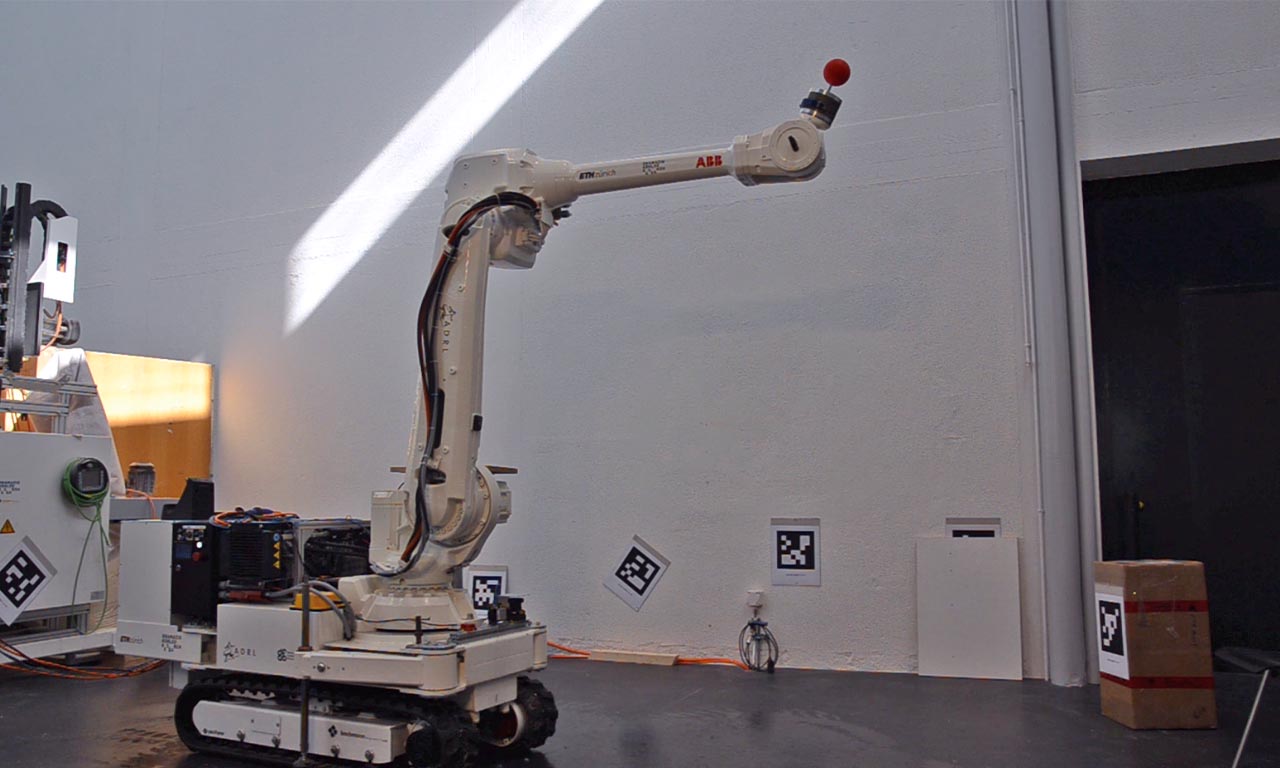} \hfill
&   \includegraphics[width = 0.196\textwidth]{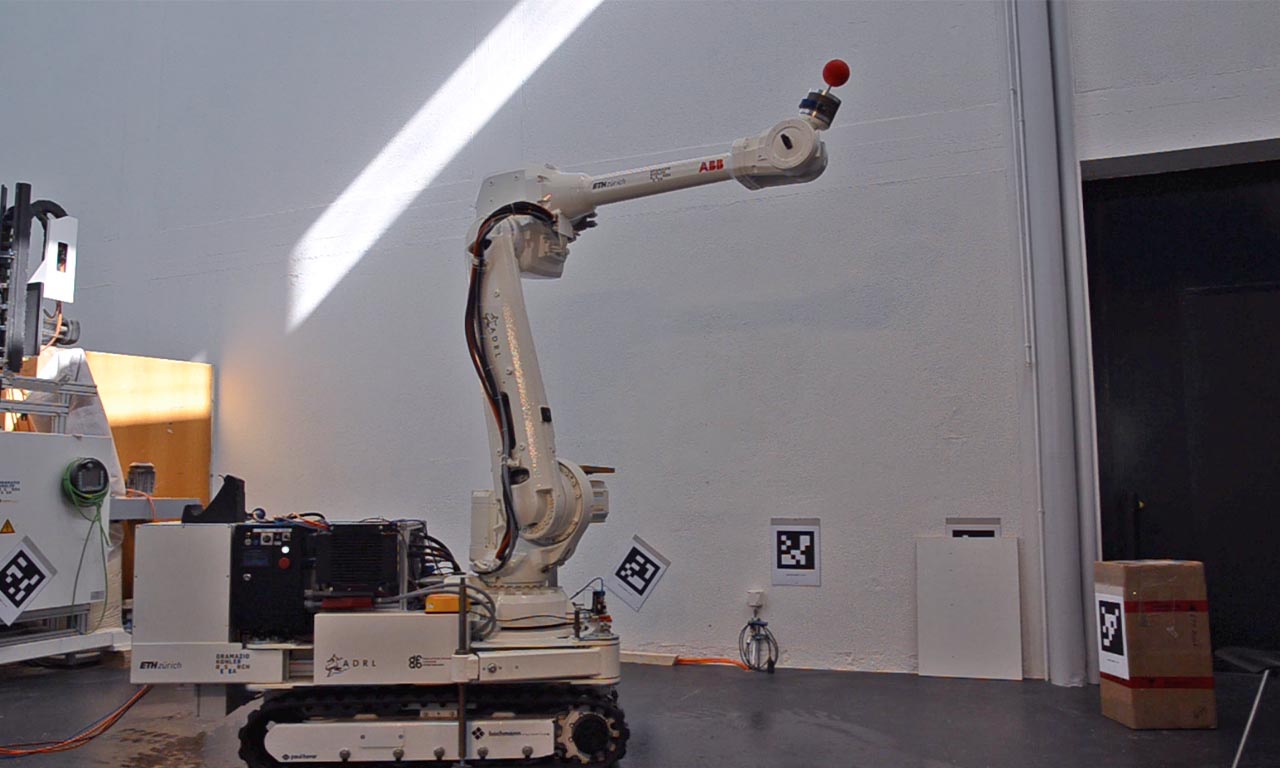} \hfill
&   \includegraphics[width = 0.196\textwidth]{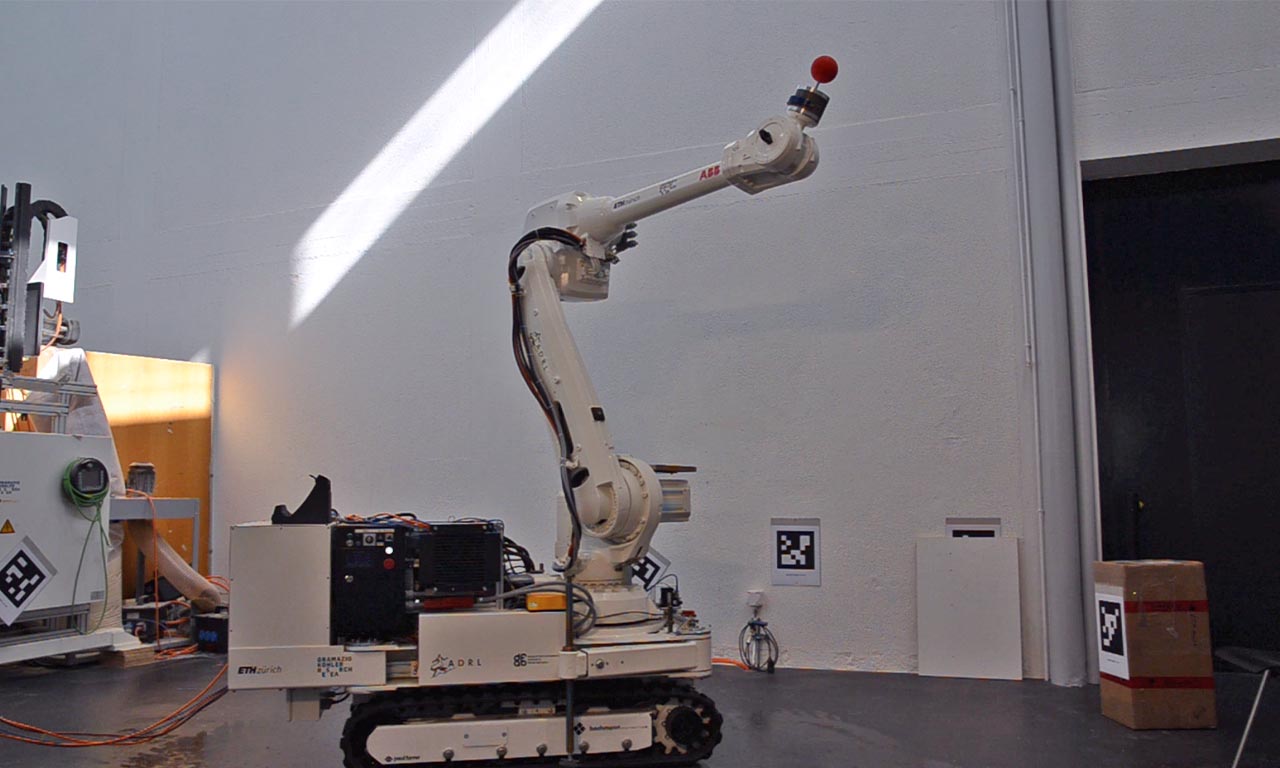} \hfill
\end{tabular}
\label{fig:IF1_experiment_sequence} 
}
\hfill 
\end{tabular} 
\renewcommand{\arraystretch}{1.0}
\caption{(a)-(b): Snapshots of motions planned with constrained SLQ for IF2. The red cylinders show the wheels, the white arrows indicate instantaneous translational velocities of the wheel centers.
(c) Motion planned for IF1 with an end effector path constraint.
(d) Snapshots from a motion sequence executed on the real IF1 using receding horizon optimal control with an end effector position constraint.}
\label{foo2}
\end{figure*}

\subsection{Implementation and simulation results}
For planning, we initialize the system with a given initial state and specify a complete full body pose (except for the wheel joint angles) as the desired terminal state $\vx_r$.
Our cost function is of the form~\eqref{eq:used_cost_function}, where $\vR$ and $\vQ_f$ are diagonal\footnote{$\vR$ is Identity except for the penalty on the translational base velocities, which was $3.0$. $\vQ_f$ is $10\cdot I$, except for the base position states, which were penalized with 100, and the wheel joint angles, which were not penalized.}.

\begin{table}[tb]
  \centering
  \caption{Constrained SLQ simulation experiments on IF1 and IF2}
  \label{tab:if2_results}
  \begin{tabular}{ccccc}
    \toprule
    Task & Time Horizon [s] & \#iter & Constr. ISE & CPU Time [s]\\
    \midrule
    IF2-A           & 12.0 & 15 & $<10^{-4}$ & 3.35  \\
    IF2-B           & 12.0 & 17 & $<10^{-3}$ & 3.92  \\
    IF1-$\infty$    & 5.0  & 8  & $<10^{-4}$ & 0.091 \\
    \bottomrule
  \end{tabular}
\end{table}
Table~\ref{tab:if2_results} summarizes main results for two example tasks, IF2-A and IF2-B. 
In task IF2-A, the robot needs to move 1.0~m in both the $x$ and $y$ directions.
In task IF2-B, the system has to translate 1.0~m away from the start and rotate its trunk about~$180^\circ$. In both tasks, it needs to adjust the wheel orientations at the maneuver start. The IF2 tasks have a 12.0~sec time horizon, were solved within less than 20 iterations and less than 4.0~sec single core CPU-time. The CPU-time value is an average over 20 independent runs. The solutions show a maximum Integrated Square Constraint Error (ISE) of less than $10^{-3}$. 
Fig.~\ref{fig:IF2_repositioning} and Fig.~\ref{fig:IF2_repositioning_and_rot} show snapshots from  visualizations of the solutions (equidistant in time).
The reader can find the videos online\footlabel{video}{\url{https://youtu.be/rVu1L_tPCoM}}.

\section{Optimal Control of a tracked Mobile Manipulator}
\label{sec:IF1_MPC}
\subsection{Modelling}
The IF1, which is shown in Fig.~\ref{fig:experimental_setup}, is a $1.5$~ton mobile manipulator with 2.55~m arm reach, which is capable of handling a 40~kg payload~\cite{sandy16}. It is an autonomous robot with integrated on-board control, sensing, and power system. The IF1's arm is a standard industrial robot arm (ABB IRB~4600). Its base is equipped with hydraulically driven tracks. With the IF1, we have previously shown digital fabrication tasks such as building a dry brick wall~\cite{doerfler2016automation}. However, a coordinated base- and end effector motion, which is important to enable new building processes, was not yet achieved. 

Since rolling and pitching of the IF1's base can typically be neglected, we describe the base kinematics in a 2D planar model. The tracks are reduced to a two-wheel model~\cite{sarkarTracks}. The state is defined as 
$\vx = [x_c \ y_c \ \theta \ \vph^\top]^\top$ and the control as 
$\vu = [\dot x_c \ \dot y_c \ \dot\theta \ \dot \vph^\top]^\top$, where $x_c$ and $y_c$ denote the world position of a frame fixed to the robot base and $\theta$ is its heading angle. As a strong simplification, we assume that the base's center of rotation (CoR) remains constant w.r.t the base frame. Hence, the non-holonomic constraint results as
$\dot y_c \cos \theta - \dot x_c \sin \theta - \dot \theta d = 0$,
where $d$ defines an offset between the CoR and the geometric center of the two-wheel model. We assume that no slip occurs, hence we can use the following equations to calculate the track speeds $v_r$, $v_l$
\begin{equation}
\begin{split}
    v_r &= \dot x_c \cos \theta + \dot y_c  \sin \theta + b \dot \theta \\
    v_l &= \dot x_c \cos \theta + \dot y_c  \sin \theta - b \dot \theta
\end{split}
\label{eq:base_state_to_track_vel}
\end{equation}
which are the real-world control inputs to the robot base.

Given the world position and velocity of the base, the end-effector position $\vr_{ee}^{fr_W}$ and velocity $\vv_{ee}^{fr_W}$ follow immediately via the forward kinematics of the arm and its Jacobian. We formulate end effector tracking constraints as either
\begin{equation}
\begin{split}
    \vg_1(\vq, \dot \vq, t) &: \quad \vv_{ee}^{fr_W} = \vv_{ee, \mathbf{ref}}^{fr_W} (t, \vq, \dot \vq) \quad \text{or} \\
    \vg_2(\vq, t) &: \quad  \vr_{ee}^{fr_W} = \vr_{ee, \mathbf{ref}}^{fr_W} (t, \vq) \ \text{.}
\end{split}
\label{eq:ee_constr_eq}
\end{equation}
The reference positions and velocities can be chosen arbitrarily as long as their first order derivatives are well-defined.

\subsection{An example for planning with end effector constraints}
\label{sec:planning_endeffector_constr}
\begin{figure}[bt]
    \centering
    \includegraphics[width=0.99\columnwidth]{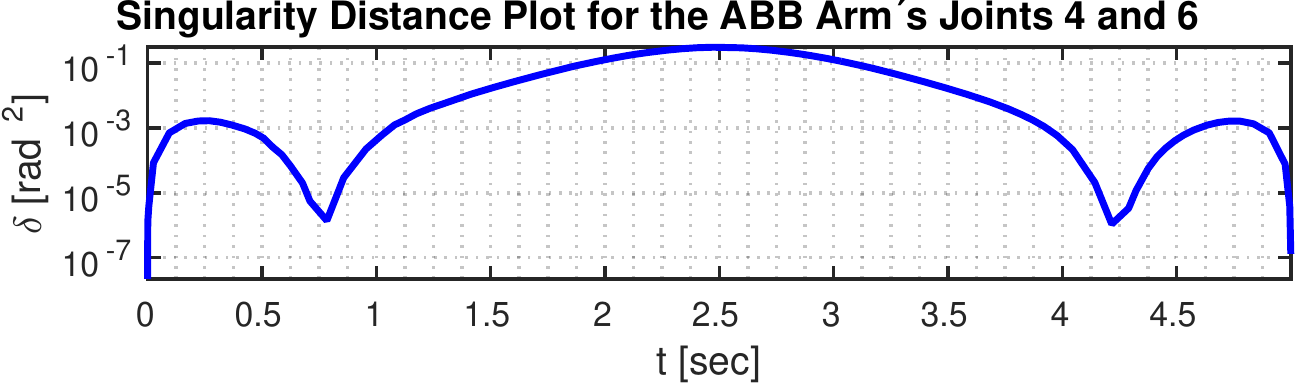}
    \caption{Distance $\delta$ from a current robot configuration to the singularity formed through joints~4 and~6 of the ABB IRB 4600. $\delta$ is defined in terms of the squared joint angle difference to the singularity. Our planning approach is able to deliberately go to and recover from singular configurations in order to fulfil the task objective and the constraints. The planned maneuver is similar to the one in Fig.~\ref{fig:IF1_figure8}, which is an operational-space tracking task.}
    \label{fig:singularity_escape}
    \vspace{1.0em}
\end{figure}

In Fig.~\ref{fig:IF1_figure8}, we show an example of a motion generated for the IF1 model, in which the end effector was constrained to follow a given task-space trajectory. The range of the motion was chosen such that the arm reach alone was insufficient to perform it without moving the base. The optimized motion satisfies the non-holonomic base constraint and the end effector tracking constraints with less than $10^{-4}$~ISE. Additional data is listed in Table~\ref{tab:if2_results}. 
It should be noted that the algorithm deliberately drives the arm into a singularity, in order to leverage the maximum reach, and also recovers from it. While many approaches that plan in the low dimensional task space are in the need of sophisticated methods to avoid singularities (e.g.~\cite{dietrich2011singularity}), our approach handles them automatically. Fig.~\ref{fig:singularity_escape} illustrates that property.

\subsection{Fast receding horizon optimal control of IF1}
In this experiment, we estimate the system state using a visual-inertial sensor, which is rigidly mounted to the base, in combination with the `Robot Centric Absolute Reference System'~\cite{neunert2016open}, which relies on fixed fiducial markers in the environment and delivers state updates at 20~Hz rate. The end effector pose is calculated from the base state and the industrial arm's joint encoder readings.

We implemented a control architecture as described in Section~\ref{sec:receding_horizon_opt}. As inner loop, we run a whole-body controller at 250~Hz.
The industrial arm is controlled via a commercial interface provided by the arm manufacturer, which allows to send joint reference positions and velocities at high rate. To control the tracks, we have implemented a custom speed tracking controller. Since the control feedback, which is calculated according to Equation~\eqref{eq:control_law_slq}, is compliant with the constraints, the updated track velocities can be obtained directly via Equation~\eqref{eq:base_state_to_track_vel}.

\begin{figure}
    \centering
    \includegraphics[width=0.98\columnwidth]{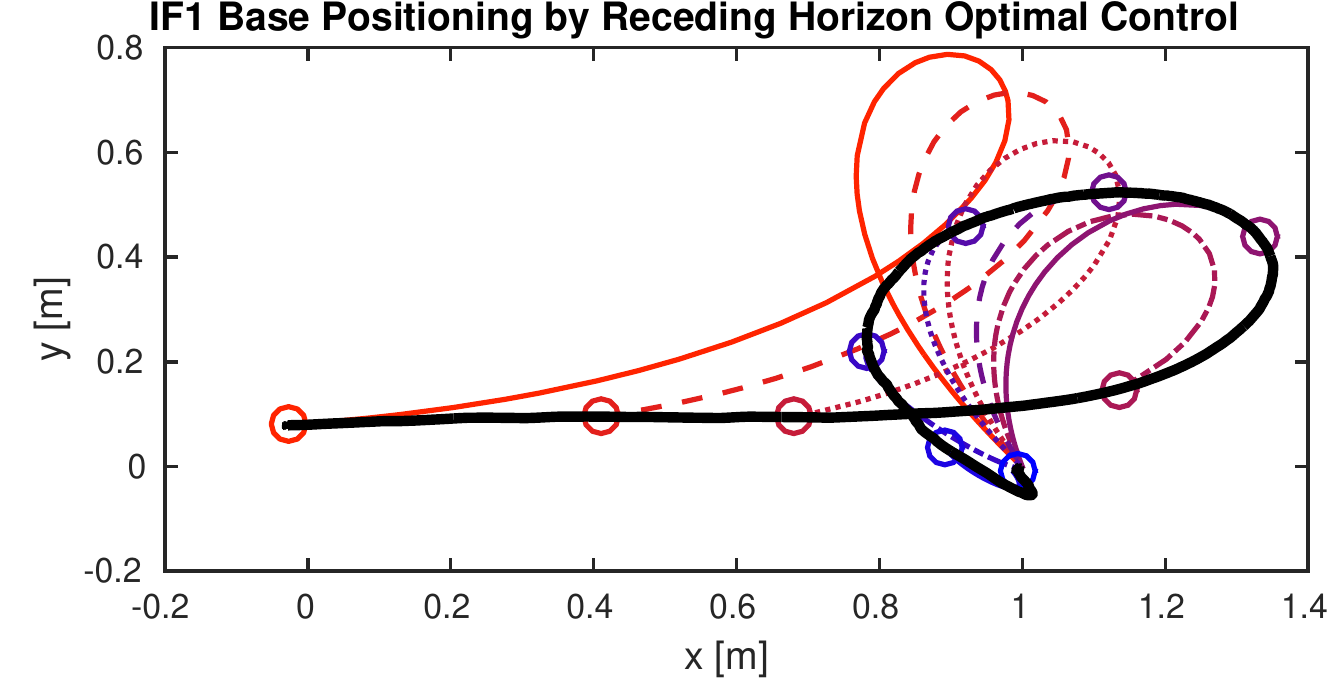}
    \caption{
    Overhead plot of continuously computed paths for a base repositioning/reorientation task using receding horizon optimal control. The initially planned path is shown in red. 
    As this motion was performed on ground with very high friction which caused the robot's rubber tracks to stick, the system started to deviate from the plan despite feedback. Hence, replanning was required. 
    Every 100th updated plan is plotted (color gradient from red to blue, circles denote positions at replanning start). The path that was actually executed by the base is shown in black.
    The controller finally drives the system to the desired location $(1,0)$ with accuracy $<2$~cm. 
    }
    \label{fig:base_replanning}
\end{figure}

Due to model uncertainty, significant slip in the tracks during driving, and various other factors, we observed that even closed-loop motions executed on the real system resulted in a large amount of end-effector constraint error and deviation of the desired final pose. Therefore, we run Constrained SLQ as a receding horizon optimal controller, as introduced in Section~\ref{sec:receding_horizon_opt}. 
We set a moving time horizon with a fixed length of 15~seconds for all scenarios.
The cost function was of the form of Equation~\eqref{eq:used_cost_function}, the weights are given below\footnote{
$\vR = \diag \left( 1, \ 1, \ 1, \ 0.1, \ \ldots, 0.1  \right)$,
$\vQ_f= \diag \left( 3, \ 3, \ 3, \ 0, \ldots, 0  \right)$}.
The capabilities of kinematic receding horizon optimal control for steering a tracked vehicle with significant uncertainties to a desired target is highlighted in Fig.~\ref{fig:base_replanning}. It shows plots of optimal paths for moving and reorienting the base from its current- to a desired pose.

Three types of experiments were performed on IF1 using the proposed control framework: 
\begin{itemize}
    \item [A]base rotation about $90^\circ$ while maintaining the initial end effector position,
    \item [B]base repositioning between 1.5~m and 3.0~m distance while maintaining the initial end effector position,
    \item [C]executing a V-shaped base motion while maintaining the end effector position.
\end{itemize}
Videos of these experiments are provided\footref{video}.
Fig.~\ref{fig:IF1_experiment_sequence} shows snapshots of task C.

\begin{figure}
    \includegraphics[width = 0.99\columnwidth]{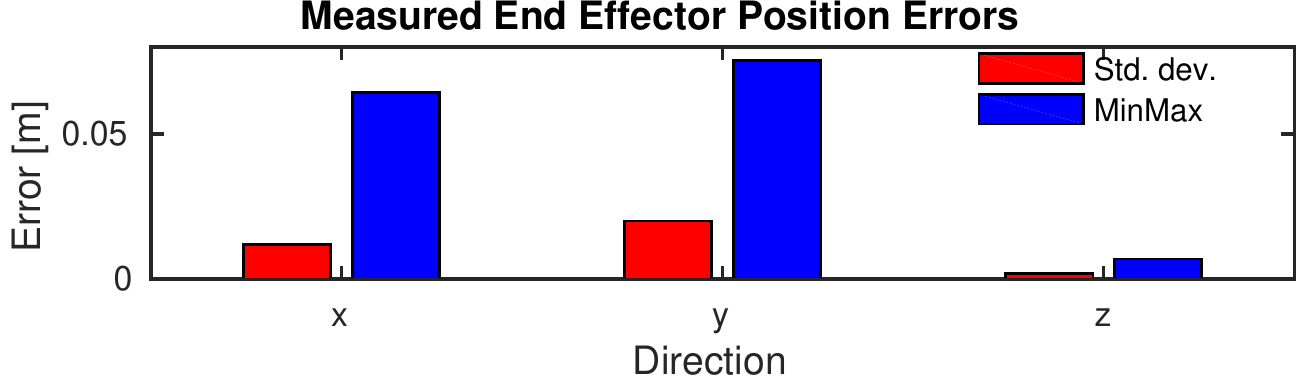}
    \caption{Min-max errors and standard deviation of the end effector position in world coordinates over the course of 10 identical experiments. The task was to keep the end effector at the same position while relocating the base. The maximum min-max end effector displacement measured was 76.4~mm.}
    \label{fig:total_station_errors}
\end{figure}

To quantify the end effector position error, we used a Hilti `Total Station' measurement system. We repeated an experiment of type B 10~times and obtained the results shown in Fig.~\ref{fig:total_station_errors}. 
Over the course of all 10 experiments, we measured a maximum min-max end effector displacement of 76.4~mm and standard deviations of 20.0~mm, 11.9~mm and 1.8~mm in $x$, $y$ and $z$ direction. There was significantly less motion in the $z$-direction, where the min-max end effector displacement was 6.8~mm and the standard deviation was 1.8~mm.

While our planner ran on a single core of an Intel Core~i7 CPU (2.30~GHz), we typically achieved replanning rates between 50~Hz and 100~Hz. In the shown experiments, convergence for the initial plan took up to 8 iterations. In warm-starting mode, convergence took 1-3 iterations.

\section{Discussion and Conclusion}
\label{sec:Discussion}
\subsection{Notes on the experiments}
In the presented experiments, we demonstrated the planning of complex repositioning maneuvers for a 26~DoF robotic base with four wheels and a total of 12~wheel motion constraints. Here, the computation time was less than a third of the time horizon of the maneuver.
Additionally, we showed the application of our algorithm to a real-world non-holonomic mobile manipulator with end effector constraints, which we controlled in a receding horizon optimal control fashion with a re-planning frequency of up to 100~Hz. Between the motion plan updates, the system was governed through constraint-compliant kinematic feedback laws designed by Constrained SLQ.
The accuracy of the end effector regularization task was limited by a number of factors:
\begin{itemize}[leftmargin=*]
\item we assumed the robot to have a fixed CoR w.r.t. the base, which is a strong assumption since the base to arm weight ratio is approximately $2.5:1$. In reality, we observed that the CoR varies significantly with the arm position. This model uncertainty could partially be compensated for by the fast update rates of the planner.
\item for safety reasons, and in order to prevent abrupt arm motion in case of a loss of communication, we had to run all experiments with a limited gain in the arm joint control, which was not accounted for in the controller design (we assumed perfect position/velocity tracking).
\item the out-of-the box base pose estimator would have required more modifications to increase its accuracy and reliability, which was impossible within the timeline of our experiments. The overall control performance suffered from occasionally occurring jumps in the base estimate at the order of centimeters. Those disturbances got directly reflected in the end effector position.
\end{itemize}

\subsection{Notes on the algorithm}
In this work, we have introduced the Constrained SLQ algorithm to the field of kinematic trajectory planning. While it is not intended to compete with existing planners in cluttered environments, it has favourable complexity properties in a local regime, i.e. linear time complexity~$O(n)$. We reach a high efficiency through a continuous-time implementation which allows the use of adaptive time-step integrators. Additionally, Constrained SLQ provides time-varying feedback laws in terms of kinematic quantities, which are locally compliant with the constraints. 

We note that, complementary to formulating equality constraints in operational space, one can also formulate operational-space goals in terms of the cost function. For example, this enables end-effector positioning for mobile manipulators subject to non-holonomic constraints\footref{video}.

\section{Outlook and Future Work}
\label{sec:Outlook}
In this work, the algorithm was running on a single CPU core. However, large parts can be parallelized, such as the linear quadratic approximation at each iteration, the controller design and the line-search. 
Therefore, there is potential for further speeding up the computations, possibly enabling receding horizon optimal control with non-holonomic constraints for even more complex systems and scenarios.

An example provided in Section~\ref{sec:planning_endeffector_constr} shows that the algorithm can leverage singularities for achieving task-space goals. However, its detailed treatment is left for future work.

In~\cite{rodriguez2010active}, an extension of the discrete-time Constrained SLQ to handle inequality constraints using an active-set approach was introduced.
In the next step, we will extend our framework to handle inequality constraints, too, which will enable the study of advanced scenarios, like obstacle avoidance, for complex robots with non-holonomic constraints. 
Also, it is desirable to benchmark the algorithm against efficient implementations of other planners.





\footnotesize
\section*{Acknowledgements}
The authors would like to thank the Hilti AG for providing a Total Station for measuring absolute robot positions, and M. Neunert for his advice regarding the RCARS system.
This research was supported by the Swiss National Science Foundation through the NCCR Digital Fabrication and a Professorship Award to J. Buchli, and a Max-Planck ETH Center for Learning Systems Ph.D. fellowship to F. Farshidian.

\bibliographystyle{ieeetr}
\bibliography{root}

\end{document}